\documentclass[journal]{IEEEtran}
\usepackage{algorithm}
\usepackage{algorithmic}

\usepackage{tabularx}
\usepackage{amsmath}
\usepackage{amssymb}
\usepackage{graphicx}
\usepackage{booktabs}  
\usepackage{threeparttable} \usepackage{multirow} 
\usepackage{color}

\usepackage{subfigure}

\newtheorem{theorem}{Theorem}
\newtheorem{proposition}{Proposition}

\newtheorem{definition}{Definition}
\newtheorem{assumption}{Assumption}

\usepackage{cite}

\ifCLASSINFOpdf
\else
\fi
\begin{document}
%
\title{Federated Soft Poly-Level Learning Over a Distributed Network with
Guaranteed Convergence}
%
%
%

\author{Yang~Jiao,~\IEEEmembership{Member,~IEEE}, Kai~Yang,~\IEEEmembership{Senior~Member,~IEEE}, Jianwei~Huang,~\IEEEmembership{Fellow,~IEEE}

}

%
%

\markboth{IEEE Transactions on Networking}%
{}
%



\maketitle

\begin{abstract}
Multi-level learning/optimization, e.g., bilevel optimization, has been recognized for its versatile applicability across a broad array of domains, spanning from signal processing to machine learning. \textcolor{black}{However, most existing works focus on bilevel and trilevel optimization. In contrast, multi-level optimization problems involving four or more hierarchical levels, which we term poly-level learning (PLL) problems, remain largely unexplored.} In addition, in practical PLL, data may be distributed across multiple nodes over a distributed network. Nevertheless, extant studies on PLL predominantly focus on the non-federated setting, necessitating the collection of data to central servers and thereby posing potential privacy risks. To this end, we propose a Federated Poly-Level Learning (FPLL) framework over a distributed network to tackle the PLL problems with soft constraints in a federated manner. FPLL hinges on a novel non-linear cut, i.e., $L$-cut, for the level-wise relaxation to the PLL problems. Additionally, the proposed federated algorithm is designed with adjustable parameters that allow for a flexible trade-off between its computational complexity and performance efficiency, enabling users to tune it according to specific application requirements.  We also carry out non-asymptotic analysis to derive the iteration complexity for the proposed FPLL to achieve the $\epsilon$-Pareto stationary point.  Extensive experiments have been undertaken to elucidate the superior efficacy of the proposed method.
\end{abstract}

\begin{IEEEkeywords}
Multi-level Optimization, Federated Learning, Soft Constraint.
\end{IEEEkeywords}

%
\IEEEpeerreviewmaketitle

\vspace{-1mm}

\section{Introduction}
\vspace{-0.5mm}

\IEEEPARstart{M}{ulti-level} learning, also called multi-level optimization, {\color{black}finds wide-ranging applications in signal processing \cite{crockett2021motivating} and machine learning \cite{choe2022betty}. For instance, multi-level optimization is employed for wireless resource allocation in dynamic environments in \cite{sun2022learning}. Likewise, in \cite{chen2024robust}, the robust beamforming problem in downlink multi-cell systems is tackled based on bilevel optimization. A bilevel optimization method, which eliminates the need for projection, is introduced for signal processing in \cite{akhtar2022projection}. In \cite{park2020learning}, the bilevel meta-learning is applied to an Internet-of-Things scenario where devices sporadically send short packets containing only a few pilot symbols. Furthermore, \cite{crockett2022bilevel} explores the use of multi-level optimization in image reconstruction. Comprehensive discussions on the application of bilevel optimization in signal processing and machine learning can be found in \cite{zhang2024introduction}.}

\textcolor{black}{Driven by recent advancements in machine learning and beyond, there is a growing trend to formulate complex decision-making processes as multi-level optimization problems with deeply nested structures. In this paper, we use poly-level learning (PLL) to refer to multi-level learning problems with four or more levels, which have been studied in many applications. For example, to fully exploit the significance of different input data elements in Neural Architecture Search (NAS), a PLL-based method is proposed in \cite{hosseini2022saliency}, which alternately optimizes two sets of model parameters, saliency maps, and the architecture according to validation performance, thereby guiding the model to focus on salient data elements while discovering superior architectures. Similarly, a PLL framework for transfer learning is proposed in \cite{xie2024simultaneous}, which alternately optimizes the domain distance network, source data selection strategy, target model parameters, and data weights of domain distance network, thereby enabling adaptive source utilization and improving target domain learning. In addition, to make augmented data more suitable for contrastive learning in Natural Language Processing (NLP), a PLL-based method is introduced in \cite{fang2022end} to jointly optimize sentence augmentation, contrastive text encoder pretraining, text classifier fine-tuning, and data reweighting, which facilitates task-aware augmentation and improves downstream text classification. Moreover, PLL is also used in meta-learning based NAS \cite{huang2025federated}, uncertainty-aware transmission network planning \cite{moreira2017five}, and joint optimization of
models and hyperparameters in machine learning pipelines \cite{jiao2022timeautoad}.} The general formulation of a PLL problem can be expressed as follows:
\begin{equation}
\label{eq:1}
\begin{array}{l}
\min \quad {f_{1}}({\boldsymbol{x}_1},{\boldsymbol{x}_2}, \cdots ,{\boldsymbol{x}_n}) \\ \vspace{1mm}
{\rm{s}}{\rm{.t}}{\rm{.   }} \, {\boldsymbol{x}_2} = \mathop {\arg \min }\limits_{\boldsymbol{x}_2'}  {{f_{2}}({\boldsymbol{x}_1},{\boldsymbol{x}_2'},\! \cdots \! ,{\boldsymbol{x}_n})} \\
\qquad \qquad \qquad \quad \; \ddots \\
\quad \quad \; \; {\rm{s.t}}{\rm{.  }} \, {\boldsymbol{x}_n} = \mathop {\arg \min }\limits_{{\boldsymbol{x}_n'}}  {{f_{n}}({\boldsymbol{x}_1},{\boldsymbol{x}_2'}, \!\cdots\! ,{\boldsymbol{x}_n'})} \\
{\mathop{\rm var}} . \qquad \qquad {\boldsymbol{x}_1},{\boldsymbol{x}_2}, \cdots ,{\boldsymbol{x}_n},
\end{array}
\end{equation}
where $n$ denotes the number of levels of the optimization problem, and $n \ge 4$ in poly-level learning. ${f_{i}}, i \in  [n]$ denotes the $i^{\rm{th}}$ level objective, and {\color{black} ${\boldsymbol{x}_i}  \in \mathbb{R}^{d_i}, i \in  [n]$} are the variables, where $[n]=\{1,\cdots,n\}$. 

\noindent\textbf{Theoretical analyses of tackling PLL problems.} \textcolor{black}{From the perspective of the polynomial hierarchy, the theoretical hardness of a multilevel optimization problem strictly increases with its nesting depth. While the decision versions of bilevel and trilevel linear optimization are $\Sigma_1^{\mathrm P}$-complete and $\Sigma_2^{\mathrm P}$-complete, respectively, the computational complexity of decision problems in $n$-level linear PLL reaches $\Sigma_{n-1}^{\mathrm P}$-completeness ($n\ge4$) \cite{sugishita2026decision}. Consequently, unless the polynomial hierarchy partially collapses (i.e., $\Delta_{n-1}^{\mathrm P} = \Sigma_{n-1}^{\mathrm P}$), higher levels represent strictly harder decision problems that admit no polynomial reduction downwards.
Practically, verification requires nested subproblem solves whose total runtime multiplies with every extra layer, leading to drastically higher computational cost.}


In PLL applications, data may be distributed across different nodes. However, existing works, e.g., \cite{choe2022betty,sato2021gradient}, primarily concentrate on addressing the PLL problems in a \textbf{non-federated} manner, which requires aggregating the data onto central servers and thus could potentially lead to issues such as the compromise of data privacy \cite{subramanya2021centralized}. In addition, in multi-level optimization, the lower-level optimization problem, that is, ${\boldsymbol{x}_n} \!=\! \mathop {\arg \min }\nolimits_{{\boldsymbol{x}_n'}}  {{f_{n}}({\boldsymbol{x}_1},{\boldsymbol{x}_2'}, \!\cdots\! ,{\boldsymbol{x}_n'})}$, often serves as a \textbf{soft constraint} \cite{jiao2025pr} to the upper-level optimization problem. This is because the solution obtained remains feasible and meaningful even if this soft constraint is partially violated. For example, in bilevel neural architecture search, the results obtained after a single gradient descent step are utilized to approximate the optimal solution of the lower-level optimization problem \cite{liu2018darts}.  In bilevel meta-learning \cite{finn2017model}, the results obtained after a few steps of gradient descent are employed as the approximation to the optimal solution in the lower-level optimization problem.  In trilevel optimization \cite{jiao2024provably}, the outcomes obtained from 
$K$ communication rounds are used as a substitute for the optimal solution of the lower-level problem in distributed trilevel learning problems.  In this work, we focus on the soft PLL problems, which refer to the PLL problems with soft lower-level constraints.


 \textcolor{black}{Moreover, multi-objective optimization has become increasingly important in machine learning, where systems frequently balance conflicting goals such as accuracy, robustness, privacy, and efficiency \cite{sener2018multi}. These trade-offs are common in domains like reinforcement learning \cite{van2014multi}, hyperparameter optimization \cite{karl2023multi}, and federated learning \cite{hu2022federated}. In multi-level learning problems, these conflicting objectives are inherently coupled with nested structures. For instance, multi-objective optimization has been integrated into bilevel frameworks to address meta-learning (with conflicting meta-level objectives optimized at the upper-level) \cite{ye2021multi}, as well as min-max settings for robust representation learning (with multiple objectives at the upper-level) \cite{gu2023min}. Beyond these two-level structures, such multi-objective trade-offs can also naturally arise in PLL paradigms. For example, meta-learning based NAS \cite{huang2025federated} may jointly optimize architectures and hyperparameters while balancing performance, model complexity, and search cost. Federated joint optimization of
models and hyperparameters in machine learning pipelines \cite{choe2022betty} may involve trade-offs among global accuracy, communication cost, privacy budget, and client-level fairness.} However, existing multi-objective optimization work has mainly focused on single-level or bilevel settings, while the PLL with upper-level multi-objectives remains less explored. Consequently, this work aims to address the following key question: \textit{How to systematically design an efficient framework to tackle soft PLL problems with multiple objectives in a federated manner?}

Additionally, cutting plane methods are extensively employed for distributed optimization, owing to their demonstrated high effectiveness and efficiency \cite{yang2008distributed,burger2013polyhedral}. Recently, several studies have explored the application of cutting plane methods for tackling multi-level learning problems in a distributed manner \cite{liu2025argus,chen2024robust,jiao2024provably}. \textcolor{black}{Compared with existing distributed optimization methods for multi-level learning, e.g., approaches based on hypergradient computation \cite{tarzanagh2022fednest,li2023communication}, gradient-tracking schemes \cite{yousefian2021bilevel}, and penalty-based reformulations \cite{niu2026hessian}, cutting plane methods offer two main advantages. First, the generated cutting planes can improve problem decomposability, enabling local workers to approximate complex feasible regions while exchanging only cutting plane parameters. Second, the number and generation frequency of cutting planes provide a flexible mechanism for balancing computational and communication complexity against solution performance. This flexibility is especially important for multi-level learning, where solving such problems is often computationally expensive.} Nevertheless, current methodologies are constrained to producing linear cutting planes for multi-level learning, thereby restricting approximation capabilities. This issue raises another key question: \textit{Is it possible to devise a non-linear cut to enhance the polynomial approximation ability in PLL?}


In addressing these questions, we propose a framework named Federated Poly-Level Learning (FPLL) as a solution to address the soft PLL problems in a federated manner. Specifically, FPLL treats the lower-level optimization problem as a \textit{soft constraint} within the framework of the upper-level problem and constructs a level-wise relaxation based on the novel non-linear cut, referred to as $L$-cut. The proposed $L$-cuts are capable of constructing a tighter relaxation than traditional methods. In addition, FPLL takes into account scenarios where conflicting objectives may arise in distributed systems and develops a distributed multi-gradient descent method. The proposed FPLL allows for flexible control over the trade-off between computational complexity and performance efficiency with adjustable parameters. Furthermore, the iteration and communication complexities for the proposed FPLL to obtain the $\epsilon$-Pareto stationary point are theoretically analyzed. 

The contributions of this work can be summarized below.

\begin{enumerate}
    \item [\textbf{1)}] 
Different from existing works, which mainly focus on addressing PLL problems in a non-federated setting, this work places particular emphasis on investigating how to effectively tackle soft PLL problems in a federated manner. Moreover, we consider a more challenging scenario in which potentially conflicting objectives may arise among different workers. 

    \item [\textbf{2)}]  An effective framework, FPLL, incorporating novel $L$-cuts, is proposed for addressing soft PLL problems in a federated manner. As far as we are aware of, this represents an early attempt to design the polynomial approximation for soft PLL problems. FPLL is designed with adjustable parameters that allow
for a flexible trade-off between its computational
complexity and performance efficiency.  Additionally, we theoretically analyze the iteration and communication complexities for FPLL to achieve the  $\epsilon$-Pareto stationary point.

 \item [\textbf{3)}] Comprehensive experiments on federated continual learning and robust fine-tuning demonstrate that FPLL consistently outperforms state-of-the-art methods, while validating the effectiveness of the proposed $L$-cuts and its communication efficiency.
\end{enumerate}

It should be noted that while this paper focuses on poly-level learning, the proposed approach is also capable of addressing bilevel and trilevel optimization problems.

\vspace{-1mm}

\section{Related Work}
\vspace{-0.5mm}
\subsection{Distributed Nested Optimization}
Recently, a large body of works has been proposed to address the nested optimization problems in a distributed manner. Broadly categorized, these works fall into two main groups: decentralized methods and centralized distributed methods. Within the decentralized methods, numerous studies, e.g.,  \cite{chen2023decentralized,yang2022decentralized,jiao2026aspire}, have been introduced to address bilevel optimization problems in a decentralized manner. In the centralized distributed method, the bilevel optimization problems under the federated setting \cite{tarzanagh2022fednest,li2023communication} are studied. In addressing distributed trilevel optimization, AFTO \cite{jiao2024provably} is introduced as a solution to the trilevel optimization problem within the federated framework. 
\textcolor{black}{Moreover, \cite{yang2023federated2} studied multi-level optimization with an arbitrary number of nested levels over decentralized networks. Different from \cite{yang2023federated2}, which considers a scalar aggregated upper-level objective, this work focuses on federated multi-level optimization with potentially conflicting upper-level objectives and develops a distributed multi-gradient method to handle such objective conflicts. }

\vspace{-1mm}
\subsection{Multi-objective Optimization}
\vspace{-0.5mm}
Multi-objective optimization is designed to tackle the challenge of concurrently minimizing multiple objectives that may inherently conflict with one another \cite{zhou2022convergence}. Multi-objective optimization algorithms can be classified into three key types: population-based \cite{rahimi2023review}, evolutionary-based \cite{alexandropoulos2019multi}, and gradient-based \cite{desideri2012multiple} approaches. Recently, gradient-based methods have garnered considerable attention due to their superior performance in solving multi-objective optimization problems, particularly within machine learning applications, e.g., multi-task learning \cite{sener2018multi}. Different from existing multi-objective works which focus on single-level (e.g., \cite{liu2021stochastic,momma2022multi}) and bilevel (e.g., \cite{ye2021multi,fernando2022mitigating,gu2023min}) optimization, the proposed framework can be applied to the multi-objective multi-level optimization problem, which is much harder to solve than multi-objective single-level and bilevel optimization problems. The comparisons between the proposed FPLL with existing works are shown in Table \ref{table:2}.



\vspace{-1mm}
\subsection{Cutting Plane Method}
\vspace{-0.5mm}
The cutting plane method is widely employed as an approximation method \cite{bertsekas2015convex,jiao2025dtzo}. The fundamental concept underlying the cutting plane method is to iteratively refine the feasible region in the optimization problem using a series of cutting planes, and this approximation will undergo progressive refinement through the successive incorporation of additional cutting planes. Cutting plane methods have found application in numerous optimization problems, including but not limited to convex optimization \cite{burger2013polyhedral,bertsekas2015convex}, robust optimization \cite{bertsimas2016reformulation,yang2014distributed} and distributed optimization \cite{yang2008distributed,burger2013polyhedral}.
Cutting plane methods have recently attracted interest in the context of distributed nested optimization. In ADBO \cite{jiao2022asynchronous}, a distributed algorithm based on the cutting plane method is proposed, which can be employed to address the distributed bilevel optimization problems. In BLRBF \cite{chen2024robust}, the cutting plane method is used to solve the distributed bilevel robust beamforming problem in downlink multi-cell systems. \textcolor{black}{In AFTO \cite{jiao2024provably}, a $\mu$-cut is introduced for distributed trilevel optimization problems, where $\mu$ denotes the weak-convexity parameter of the underlying $\mu$-weakly convex function.} The $\mu$-cut has broader applicability than cutting planes in \cite{jiao2022asynchronous,chen2024robust} since it can be applied to non-convex optimization, specifically $\mu$-weakly convex problems. Nevertheless, the current cutting plane methods for distributed multi-level optimization can only generate the linear cutting plane. In this work, a \textit{non-linear cut}, referred to as $L$-cut, is proposed for PLL problems, which is capable of constructing a tighter polynomial relaxation. Table \ref{table:cp} compares the proposed $L$-cut against traditional cutting plane methods.

\begin{table}[t]
\color{black}
\caption{Comparison between the proposed FPLL with state-of-the-art methods.}
\renewcommand\arraystretch{1.1}
\renewcommand\tabcolsep{2pt}
\centering
\scalebox{1}{\begin{tabular}{l|ccccc} 
\toprule
Method  & Bilevel  & Trilevel & Poly-level & Federated  &  Multi-objective
\\
\hline
 ADBO \cite{jiao2022asynchronous} &  \checkmark  &    &    &  \checkmark &    \\
 FEDNEST \cite{tarzanagh2022fednest} &  \checkmark  &    &   & \checkmark  &    \\
 AFTO \cite{jiao2024provably} &  \checkmark  &  \checkmark  &   & \checkmark &      \\
 FMGDA \cite{yang2023federated} &    &   &    & \checkmark  & \checkmark     \\
MORBiT \cite{gu2023min} & \checkmark   &   &    &   & \checkmark     \\
\hline
\textbf{FPLL}   &  \checkmark &  \checkmark &  \checkmark  &  \checkmark  &  \checkmark \\
\bottomrule
\end{tabular}}
\label{table:2}
\end{table}

\section{Federated Soft Poly-level Learning}

In the applications of PLL, data might be distributed over several nodes. Collecting data from nodes to central servers may lead to risks to data privacy \cite{subramanya2021centralized}. Thus, the development of an effective framework capable of tackling the PLL problems in a federated manner becomes imperative. In addition, a more complicated PLL scenario is considered in this work, i.e., there are instances wherein potentially conflicting objectives coexist within a distributed system \cite{yang2023federated,hu2022federated}, which can be expressed as a federated PLL problem below:
\begin{equation}
\label{eq:dis-MMO}
    \begin{array}{l}
\min f = {\left( {{f_{1,1}}({\boldsymbol{x}_1}, \cdots ,{\boldsymbol{x}_n}), \cdots ,{f_{1,M}}({\boldsymbol{x}_1}, \cdots ,{\boldsymbol{x}_n})} \right)^{\top}}\\
{\rm{s}}{\rm{.t}}{\rm{.   }} \; {\boldsymbol{x}_2} = \mathop {\arg \min }\limits_{{\boldsymbol{x}_2'}} \sum\limits_{j \in [M]} {{f_{2,j}}({\boldsymbol{x}_1}, {\boldsymbol{x}_2'}, \cdots ,{\boldsymbol{x}_n})} \\
\qquad \qquad \qquad \qquad \quad \ddots \\
\qquad \quad  {\rm{s.t}}{\rm{.  }} \; {\boldsymbol{x}_n} = \mathop {\arg \min }\limits_{{\boldsymbol{x}_n'}} \sum\limits_{j \in [M]} {{f_{n,j}}({\boldsymbol{x}_1},{\boldsymbol{x}_2'}, \cdots ,{\boldsymbol{x}_n'})} \\
{\mathop{\rm var}} . \qquad \qquad {\boldsymbol{x}_1},{\boldsymbol{x}_2}, \cdots ,{\boldsymbol{x}_n},
\end{array}
\end{equation}
where $f_{i,j}$ represents the local $i^{\rm{th}}$-level objective at the $j^{\rm{th}}$ worker, $\forall i,j$, which is smooth, and $\boldsymbol{x}_{i} \in \mathbb{R}^{d_i}, \forall i \in [n]$ denote the optimization variables. {\color{black} In particular, $f_{1,j}$ represents the local upper-level objective at the $j^{\rm{th}}$ worker, and hence the upper-level objective is also distributed across the workers. The local upper-level objectives are retained separately to explicitly account for potential conflicts among different workers.} {\color{black} In this work, we adopt a parameter-server architecture, i.e., a star topology consisting of a central server and $M$ workers.} This work focuses on the federated soft PLL problems, i.e., the lower-level problem in Eq. (\ref{eq:dis-MMO}) acts as a \textit{soft constraint} to the upper-level optimization problem. {\color{black}Specifically, a soft constraint allows the lower-level optimality condition to be relaxed, such that it is satisfied approximately rather than exactly. For example, the $n^{\rm{th}}$ level optimization problem ${\boldsymbol{x}_n} = \mathop {\arg \min }\nolimits_{{\boldsymbol{x}_n'}} \sum\nolimits_{j \in [M]} {{f_{n,j}}({\boldsymbol{x}_1}, \cdots ,{\boldsymbol{x}_n'})}$ can be relaxed to $||{\boldsymbol{x}_n}-  \mathop {\arg \min }\nolimits_{{\boldsymbol{x}_n'}} \sum\nolimits_{j \in [M]} {{f_{n,j}}({\boldsymbol{x}_1}, \cdots ,{\boldsymbol{x}_n'})}||^2 \le \varepsilon_n$, where $\varepsilon_n \ge 0$ represents the tolerance.} The key challenges encountered in tackling the federated soft PLL problem are twofold: 1) the PLL problem exhibits a high degree of nesting, and 2) the objectives may potentially conflict with each other.

\noindent\textcolor{black}{\textbf{Overview of the proposed framework.}
The proposed FPLL framework consists of three main components. First, in Sec. \ref{sec:level-wise}, we construct a level-wise $L$-cut relaxation for the federated soft PLL problem. The key idea is to treat lower-level optimization problems as soft constraints and approximate the corresponding feasible regions using level-wise $L$-cuts. Second, in Sec. \ref{sec:refine}, we present how to generate and update the $L$-cuts to progressively refine the relaxation, including adding new cuts to tighten the feasible region and removing inactive cuts to control the computational and communication overhead. Finally, in Sec. \ref{sec:distributed}, we develop a distributed multi-gradient descent method to solve the resulting problem with potentially conflicting objectives.} 

Additionally, the proposed FPLL framework can be applied to a broad spectrum of optimization problems with minor modifications. For example, for PLL problems without conflicting objectives, the distributed multi-gradient descent method can be replaced by distributed gradient descent. For trilevel optimization problems, we can set $n=3$ in the proposed framework.

\vspace{-2mm}

\subsection{Level-wise $L$-cut Relaxation}
\label{sec:level-wise}

\begin{table}[t]
\color{black}
\caption{Comparison between the proposed $L$-cut with state-of-the-art cutting plane methods.}
\renewcommand\arraystretch{1.2}
\renewcommand\tabcolsep{7pt}
\centering
\scalebox{1.0}{\begin{tabular}{c|cccc} 
\toprule
Method  & Convex  & Non-convex & Non-linear   &  Poly-level
\\
\hline

\cite{jiao2022asynchronous} &  {\checkmark}   &   &   &  \\
\cite{chen2024robust} &  {\checkmark}   &   &   &   \\
\cite{jian2024tri} &  {\checkmark}   &   &   &   \\
\cite{liu2025argus} &  {\checkmark}   &    &   &   \\
\cite{jiao2024provably} &  {\checkmark}   &  {\checkmark}  &   &   \\
\hline
\textbf{$L$-cut}   &  {\checkmark}  &  {\checkmark}  &  {\checkmark}   &  {\checkmark}  \\
\bottomrule
\end{tabular}}
\label{table:cp}
\end{table}

Cutting plane methods are extensively employed for addressing nested optimization problems in a distributed manner. As demonstrated in \cite{jiao2024provably}, developing distributed algorithms for nested optimization problems is \textit{significantly facilitated} by utilizing cutting plane methods.  \textcolor{black}{Building upon previous work on smoothing optimization \cite{nesterov2005smooth}, we introduce \emph{$L$-cuts} for functions with Lipschitz continuous gradients, where $L$ denotes the gradient Lipschitz constant, to approximate the feasible region induced by the corresponding constraint.} We demonstrate in Propositions \ref{prop:1} and \ref{prop:2} that the proposed $L$-cuts are able to establish a relaxation of the original feasible region, and the relaxation will be progressively tightened with $L$-cuts incorporated. In this section, how to construct the level-wise $L$-cut relaxation will be introduced, details about generating the proposed $L$-cuts will be provided in Sec. \ref{sec:refine}.

\textcolor{black}{We first illustrate the construction of $L$-cut relaxation at the $n^{\rm th}$ level.} In the federated soft PLL problem in Eq. (\ref{eq:dis-MMO}), the $n^{\rm{th}}$ level problem is regarded as the constraint to the $(n-1)^{\rm{th}}$ level optimization problem, and it is equivalent to the constraint ${g_n}({\boldsymbol{x}_1}, \cdots ,{\boldsymbol{x}_n}) = 0$, where ${g_n}({\boldsymbol{x}_1}, \cdots ,{\boldsymbol{x}_n}) = ||{\boldsymbol{x}_n} - \mathop {\arg \min }\nolimits_{{\boldsymbol{x}_n'} } \sum\nolimits_{j \in [M]} {{f_{n,j}}({\boldsymbol{x}_1}, \cdots ,{\boldsymbol{x}_n'})} |{|^2}$. Solving a distributed optimization problem becomes necessary when an exact solution to $\mathop {\arg \min }\nolimits_{{\boldsymbol{x}_n'} } \sum\nolimits_{j \in [M]} {{f_{n,j}}({\boldsymbol{x}_1}, \cdots ,{\boldsymbol{x}_n'})}$ is required. Combining the perspective that the lower-level optimization problem acts as a \textit{soft constraint} for the upper-level problem in federated soft PLL, with the established practice of using $Q$-step gradient descent to estimate lower-level solutions \cite{ji2021bilevel,sato2021gradient,choe2022betty,jiao2026first}, we approximate $\mathop {\arg \min }\nolimits_{{\boldsymbol{x}_n'} } \sum\nolimits_{j \in [M]} {{f_{n,j}}({\boldsymbol{x}_1}, \cdots ,{\boldsymbol{x}_n'})}$ using the outcomes obtained after $Q$ steps, i.e.,  ${g_n}({\boldsymbol{x}_1}, \cdots ,{\boldsymbol{x}_n})
 \!=\! || \boldsymbol{x}_n - \boldsymbol{x}_n^0 + \frac{\eta_{\boldsymbol{x}_n}}{M} \!  \sum\nolimits_{j \in [M]} \!  \Delta_{n,j}  ||^2$, where $\Delta_{n,j}=\sum\nolimits_{q=0}^{Q-1}
\nabla_{\boldsymbol{x}_n}
{{f_{n,j}}({\boldsymbol{x}_1}, \cdots ,{\boldsymbol{x}_{n,j}^q})}$ denotes the accumulated local gradient obtained by worker $j$ after $Q$ gradient descent steps.

 
 Inspired by the cutting plane methods \cite{bertsekas2015convex,bertsimas2016reformulation,jiao2022asynchronous,jiao2024provably}, the $n^{\rm{th}}$ level $L$-cuts are used to build a polynomial relaxation of the feasible region concerning the constraints ${g_n}({\boldsymbol{x}_1}, \cdots ,{\boldsymbol{x}_n})=0$. Specifically, the $n^{\rm{th}}$ level feasible region $\boldsymbol{\mathcal{Q}}^t_{n}$ formed by $n^{\rm{th}}$ level $L$-cuts in $t^{\rm{th}}$ iteration can be expressed as
\begin{equation}
\label{eq:12_20_7}
\begin{array}{l}
\!\!\boldsymbol{\mathcal{Q}}^t_{n} \!=\!   \{\! \sum\limits_{r \in [n]} \! \sum\limits_{k \in [d_r]} \! \! { {a_{r, k, l}^{(n)} x_{r,k}^2} \!+\! {b_{r, k, l}^{(n)} x_{r,k}} }\! -c_l^{(n)} \!\le\! \varepsilon_n, l \! \in \! \left[|\boldsymbol{\mathcal{Q}}^t_{n}|\right]\! \},
\end{array} 
\end{equation}
where $|\boldsymbol{\mathcal{Q}}^t_{n}|$ represents the number of $L$-cuts in $\boldsymbol{\mathcal{Q}}^t_{n}$, $x_{r,k} \! \in \! \mathbb{R}^1$ represents the $k^{\rm{th}}$ element in $\boldsymbol{x}_r, r \!\in \! [n]$, $d_r$ is the dimension of $\boldsymbol{x}_r$, and  $a_{r, k, l}^{(n)} \! \in \! \mathbb{R}^1, b_{r, k, l}^{(n)} \! \in \! \mathbb{R}^1, c_l^{(n)} \! \in \! \mathbb{R}^1$ are the parameters of the $l^{\rm{th}}$ $L$-cut in $\boldsymbol{\mathcal{Q}}_{n}^t$, $\varepsilon_n \ge 0$ is a constant. \textcolor{black}{The detailed derivation of the $L$-cut parameters is provided in 
Appendix~\ref{appendix:L_cut}.} It is seen from Eq. (\ref{eq:12_20_7}) that the feasible region formed by linear cutting planes, i.e., polytope \cite{boyd2004convex}, is the special case of $\boldsymbol{\mathcal{Q}}^t_{n}$ when setting $a_{r, k, l}^{(n)}=0, \forall r, k$. It is demonstrated in Proposition \ref{prop:1} that $\boldsymbol{\mathcal{Q}}^t_{n}$ is a relaxation of the feasible region of constraint ${g_n}({\boldsymbol{x}_1}, \cdots ,{\boldsymbol{x}_n})=0$, and the relaxation will be incrementally tightened by adding 
$L$-cuts.  By combining with the $n^{\rm{th}}$ level $\boldsymbol{\mathcal{Q}}_{n}^t$ in Eq. (\ref{eq:12_20_7}), the obtained optimization problem can be expressed as
\begin{equation}
\label{eq:12-8-layern}
    \begin{array}{l}
\min  {\left( {{f_{1,1}}({\boldsymbol{x}_1}, \cdots ,{\boldsymbol{x}_n}), \cdots ,{f_{1,M}}({\boldsymbol{x}_1}, \cdots ,{\boldsymbol{x}_n})} \right)^{\top}}\\ \vspace{1mm}
{\rm{s}}{\rm{.t}}{\rm{.   }} \; {\boldsymbol{x}_2} = \mathop {\arg \min }\limits_{{\boldsymbol{x}_2'}} \sum\limits_{j \in [M]} {{f_{2,j}}({\boldsymbol{x}_1},{\boldsymbol{x}_2'}, \cdots ,{\boldsymbol{x}_n})} \\
\qquad \qquad \qquad \qquad  \ddots \\
\;\;  {\rm{s.t}}{\rm{.  }} \; \sum\limits_{r \in [n]} \! \sum\limits_{k \in [d_r]} \! { {a_{r, k, l}^{(n)} x_{r,k}^2} \!+\! {b_{r, k, l}^{(n)} x_{r,k}} } -c_l^{(n)} \le \varepsilon_n , l \! \in \! \left[|\boldsymbol{\mathcal{Q}}_{n}^t|\right]  \\
{\mathop{\rm var}} . \qquad \qquad \qquad {\boldsymbol{x}_1},{\boldsymbol{x}_2}, \cdots ,{\boldsymbol{x}_n}. 
\end{array}
\end{equation}

\textcolor{black}{The same relaxation idea can then be applied recursively from the bottom level to the upper levels. After the $n^{\rm th}$ level constraint is relaxed by $\boldsymbol{\mathcal{Q}}_n^t$, the $(n-1)^{\rm th}$ level problem can be viewed as a soft constraint, which is equivalent to constraint ${g_{n - 1}}({\boldsymbol{x}_1}, \cdots ,{\boldsymbol{x}_n}) = 0$, where ${g_{n - 1}}({\boldsymbol{x}_1}, \cdots ,{\boldsymbol{x}_n})  = 
 ||    
{\boldsymbol{x}_{n - 1}}  - \mathop {\arg \min }\nolimits_{{\boldsymbol{x}_{n - 1}'}}  \{  \sum\nolimits_{j}  {{f_{n-1,j}}({\boldsymbol{x}_1}, \cdots ,{\boldsymbol{x}_{n - 1}'},{\boldsymbol{x}_n})}, {\rm{s.t.}} \, {\boldsymbol{x}_1},  \cdots  , {\boldsymbol{x}_n} \in  \boldsymbol{\mathcal{Q}}_{n}^t \} |{|^2} $.}  Consequently, by conducting level-wise $L$-cut relaxation, we can obtain the resulting optimization problem:
\begin{equation}
\label{eq:result_L_cut_problem}
    \begin{array}{l}
\min  {\left( {{f_{1,1}}({\boldsymbol{x}_1}, \cdots ,{\boldsymbol{x}_n}), \cdots ,{f_{1,M}}({\boldsymbol{x}_1}, \cdots ,{\boldsymbol{x}_n})} \right)^{\top}} \vspace{2mm}\\
{\rm{s.t}}{\rm{.}} \, \sum\limits_{r \in [n]} \! \sum\limits_{k \in [d_r]} \! { {a_{r, k, l}^{(2)} x_{r,k}^2} \!+\! {b_{r, k, l}^{(2)} x_{r,k}} } -c_l^{(2)} \le \varepsilon_2, l \! \in \! \left[|\boldsymbol{\mathcal{Q}}^t_{2}|\right]\\
{\mathop{\rm var}} .  \qquad \qquad \qquad {\boldsymbol{x}_1},{\boldsymbol{x}_2}, \cdots ,{\boldsymbol{x}_n}.
\end{array}
\end{equation}
where $a_{r, k, l}^{(2)}\! \in \! \mathbb{R}^1, b_{r, k, l}^{(2)}\! \in \! \mathbb{R}^1, c_l^{(2)}\! \in \! \mathbb{R}^1$ denote the parameters of the $l^{\rm{th}}$ $L$-cut in $\boldsymbol{\mathcal{Q}}^t_{2}$.

\begin{table*}[t]
\color{black}
\caption{Comparison with classical federated learning formulations in terms of convergence rates.}
\renewcommand\arraystretch{1.5}
\renewcommand\tabcolsep{15pt}
\centering
\scalebox{1.1}{\begin{tabular}{l|cc} 
\toprule
Method  & Iteration complexity  & Communication complexity 
\\
\hline

Single-level single-objective \cite{haddadpour2019local} &  $\mathcal{O}(\frac{1}{\epsilon})$   &  $\mathcal{O}(\frac{d_1 M}{\sqrt[3]{\epsilon}})$ \\
Single-level multi-objective \cite{hu2022federated} &  $\mathcal{O}(\frac{1}{\epsilon})$   &  \textemdash     \\
Bilevel single-objective \cite{liu2025argus} &  $\mathcal{O}(\frac{1}{\epsilon})$   &  $\mathcal{O}(\frac{(d_1+d_2)M}{\epsilon}  + \frac{(d_1+d_2)MK_1}{\tau})$     \\
Trilevel single-objective \cite{jiao2024provably} &  $\mathcal{O}(\frac{1}{\epsilon^2})$   &   $\mathcal{O}(\frac{\sum\nolimits_{i=1}^3 d_i M}{\epsilon^2}  + \frac{\sum\nolimits_{i=1}^3 d_i M K_1}{\tau})$     \\
\hline
\textbf{Multi-level multi-objective (this work)}   &    $\mathcal{O}(\frac{1}{\epsilon})$  & $\mathcal{O}(\frac{\sum\nolimits_{i=1}^n  d_i M}{\epsilon}  + \frac{\sum\nolimits_{i=1}^n d_i M K_1}{\tau})$         \\
\bottomrule
\end{tabular}}
\label{table:classic}
\begin{flushleft}
\footnotesize
Here, \textemdash{} indicates that the corresponding result was not reported in the cited work. $M$ denotes the number of workers in the distributed system, and $\epsilon$ specifies the target accuracy, corresponding to an $\epsilon$-stationary point in single-objective optimization and an $\epsilon$-Pareto stationary point in multi-objective optimization. $K_1$ and $\tau$ represent the number of refinement-phase iterations and the update frequency for the cutting planes, respectively. Lastly, $d_i$ indicates the dimensionality of the optimization variable at the $i^{\rm{th}}$ level. 
\end{flushleft}
\end{table*}



\subsection{Refining Level-wise $L$-cut Relaxation}
\label{sec:refine}

In this section, we discuss how to update $L$-cuts to progressively refine the proposed level-wise $L$-cut relaxation. \textcolor{black}{The $L$-cuts are updated once every $\tau$ master iterations during the first $K_1$ iterations, where $\tau$ controls the update frequency of the $L$-cuts. We consider $1 \le \tau \le K_1$ such that the level-wise $L$-cut relaxation is refined at least once during this refinement phase.} Here, $K_1$ is a constant that determines the duration of the $L$-cut refinement phase and can be flexibly adjusted based on the computational and communication capabilities within the distributed system, as discussed in Theorem \ref{theorem:1}.

The level-to-level updates are carried out sequentially, progressing from the $n^{\rm{th}}$ level $L$-cut to the second level $L$-cut. Specifically, the updates for the  $i^{\rm{th}}$ level $L$-cut involve two distinct steps: 1) generation of new $i^{\rm{th}}$ level $L$-cut, and 2) removal of inactive $i^{\rm{th}}$ level $L$-cuts.

\subsubsection{Generating new $i^{\rm{th}}$ level $L$-cut}

Following \cite{ji2021bilevel,yang2021provably}, we first give the definition of $L$-smoothness.
\begin{definition}
\label{def:L-smooth}
    \textbf{($L$-smoothness/Gradient Lipschitz)} A function $g_i(\cdot)$ is $L$-smooth if for any $({\boldsymbol{x}_1}, \cdots ,{\boldsymbol{x}_n})$ and $({\hat{\boldsymbol{x}}_1}, \cdots ,{\hat{\boldsymbol{x}}_n})$ there exists a constant $L (L>0)$ satisfying
\begin{equation}
\label{eq:l_smooth}
 \begin{array}{l}
 || \nabla g_i({\boldsymbol{x}_1}, \cdots ,{\boldsymbol{x}_n}) - \nabla g_i({\hat{\boldsymbol{x}}_1}, \cdots ,{\hat{\boldsymbol{x}}_n})||  \\ \le L || [\boldsymbol{x}_1 - \hat{\boldsymbol{x}}_1;\cdots,\boldsymbol{x}_n - \hat{\boldsymbol{x}}_n ]||.
\end{array}
\end{equation}
\end{definition}

The $i^{\rm{th}}$ level $L$-cut is generated based on the property of $L$-smoothness in Definition \ref{def:L-smooth} and the point in $({t+1})^{\rm{th}}$ iteration  $({\boldsymbol{x}_1^{t+1}}, \cdots,{\boldsymbol{x}_n^{t+1}})$. Specifically, the new $i^{\rm{th}}$ level $L$-cut $\mathcal{C}_i^{\rm{new}}$ can be generated as
\begin{equation}
\label{eq:L_cut_generate}
\begin{array}{l}
\! {\left[ \! \begin{array}{l}
\! \frac{{\partial {g_i}(\boldsymbol{x}_1^{t + 1}, \cdots   ,\boldsymbol{x}_n^{t + 1})}}{{\partial {\boldsymbol{x}_1}}}\\
\qquad \quad \; \vdots \\
\! \frac{{\partial {g_i}(\boldsymbol{x}_1^{t + 1}, \cdots  ,\boldsymbol{x}_n^{t + 1})}}{{\partial {\boldsymbol{x}_n}}}
\end{array} \! \right]^{\top}} \! \left[\! \begin{array}{l}
{\boldsymbol{x}_1} \! - \!\boldsymbol{x}_1^{t + 1}\\
\quad \; \;  \vdots \\
{\boldsymbol{x}_n} \!-\! \boldsymbol{x}_n^{t + 1}
\end{array} \! \right] \!+\! g_i(\boldsymbol{x}_1^{t + 1}, \! \cdots \! ,\boldsymbol{x}_n^{t + 1})
\vspace{1mm}\\ \le \frac{L}{2}\sum\nolimits_{i \in [n]} {||{\boldsymbol{x}_i} - \boldsymbol{x}_i^{t + 1}|{|^2}} + \varepsilon_i.
\end{array}
\end{equation}

In Propositions \ref{prop:1} and \ref{prop:2}, we demonstrate that the generated $L$-cuts are capable of constructing the relaxation for the original feasible region, and the relaxation will tighten progressively as $L$-cuts are incorporated.  Subsequently, the newly generated $i^{\rm{th}}$ level $L$-cut will be added in $i^{\rm{th}}$ level feasible region as
\begin{equation}
\label{eq:2_2_17}
    \boldsymbol{\mathcal{Q}}_i^{t+1} = \boldsymbol{\mathcal{Q}}_i^{t} \cup \mathcal{C}_i^{\rm{new}}.
\end{equation}

\subsubsection{Removing inactive $i^{\rm{th}}$ level $L$-cut}
An abundance of cutting planes may result in a considerable increase in the volume of additional message passing \cite{yang2014distributed,jiao2022distributed}. To enhance the efficiency of the proposed framework, we consider removing the inactive $i^{\rm{th}}$ level $L$-cuts as follows:
\begin{equation}
\label{eq:1_17_19}
\boldsymbol{\mathcal{Q}}_i^{t + 1}\!\! = \!  \left\{\!\!\! \begin{array}{l}
\! \boldsymbol{\mathcal{Q}}_i^{t + 1} \!\setminus\! \mathcal{C}_i^{l}, {\rm{if}}  \sum\limits_{r \in [n]} \! \sum\limits_{k \in [d_r]} \!\! { {a_{r, k, l}^{(i)} x_{r,k}^2} \!+\! {b_{r, k, l}^{(i)} x_{r,k}} } \!-\! c_l^{(i)} \!\!<\! \varepsilon_i,  \\
\! \boldsymbol{\mathcal{Q}}_i^{t + 1}, {\rm{otherwise}},
\end{array} \right.
\end{equation}
where $\boldsymbol{\mathcal{Q}}_i^{t + 1} \setminus \mathcal{C}_i^{l}$ denotes that $L$-cut $\mathcal{C}_i^{l}$ is removed from $\boldsymbol{\mathcal{Q}}_i^{t + 1}$. The proposed framework is outlined in detail in Algorithm \ref{algorithm:fmmo}.

\begin{algorithm}[t]
   \caption{FPLL: Federated Poly-Level Learning}
\begin{algorithmic}
   \STATE {\bfseries Initialization:}  master iteration $t = 0$, variables ${{\boldsymbol{x}}_{1}^0}$, $ {{\boldsymbol{x}}_{2}^0}, \cdots, {{\boldsymbol{x}}_{n}^0}$ .
   
   \REPEAT
   
   \FOR{\emph{local worker}}
   \STATE computes the local gradient  $\nabla {q_{1,j}}(\boldsymbol{x}_1^t, \cdots ,\boldsymbol{x}_n^t)$ and transmit the gradient to the master;
   \ENDFOR
   
   \FOR{\emph{master}}
   \STATE computes $\boldsymbol{p}^*$ by solving problem in Eq. (\ref{eq:12_8_eq6}) and computes search direction ${\boldsymbol{S}^{t + 1}} $ by Eq. (\ref{eq:1_17_16});

   \STATE updates variables $\boldsymbol{x}_i^{t + 1}, i \in [n]$ by Eq. (\ref{eq:1_17_17}) and broadcasts the updated variables to the workers;
   \ENDFOR

   \IF{$t<K_1$ and $(t+1)$ mod $\tau$ $==0$}
   \FOR{$i=n,n-1,\cdots,2$}

   \STATE   new $i^{\rm{th}}$ level $L$-cut $\mathcal{C}_i^{\rm{new}}$ is generated by Eq. (\ref{eq:L_cut_generate}) and added by Eq. (\ref{eq:2_2_17});

   \STATE inactive $i^{\rm{th}}$ level $L$-cuts are removed  by Eq. (\ref{eq:1_17_19});
   
   \ENDFOR
   \ENDIF

   \STATE $t =t+1$;
   \UNTIL{termination.}

\end{algorithmic}
\label{algorithm:fmmo}
\end{algorithm}

\subsection{Distributed Multi-gradient Descent}
\label{sec:distributed}

In poly-level learning problems, various factors, such as Non-IID data in each node \cite{hu2022federated,mohri2019agnostic}, contribute to potentially conflicting objectives within the distributed system.
In this section, we formulate a distributed multi-gradient descent method designed to mitigate the impact of potentially conflicting objectives in PLL problems. {\color{black}Within the developed approach, we compute a search direction $\boldsymbol{S}$ at each iteration that aims to converge to a Pareto stationary point by mitigating conflicts among multiple gradients.} Furthermore, in order to protect data privacy, only the model parameters and gradients are communicated between the master and the workers. First, inspired by the penalty method in multi-level optimization \cite{lu2024first}, the constrained optimization problem in Eq. (\ref{eq:result_L_cut_problem}) is converted into an unconstrained optimization problem, as shown below:
\begin{equation}
\label{eq:4_28_13}
\begin{array}{l}
\min {\left( {{q_{1,1}}({\boldsymbol{x}_1}, \cdots ,{\boldsymbol{x}_n}), \cdots ,{q_{1,M}}({\boldsymbol{x}_1}, \cdots ,{\boldsymbol{x}_n})} \right)^{\top}}\\
{\mathop{\rm var}} .  \qquad \qquad{\boldsymbol{x}_1},{\boldsymbol{x}_2}, \cdots ,{\boldsymbol{x}_n},
\end{array}
\end{equation}
where function ${q_{1,j}}({\boldsymbol{x}_1}, \cdots ,{\boldsymbol{x}_n}) = {f_{1,j}}({\boldsymbol{x}_1}, \cdots ,{\boldsymbol{x}_n}) + \sum_l \lambda_{1,l} [\max\{h_l({\boldsymbol{x}_1}, \cdots ,{\boldsymbol{x}_n}) , 0\}]^2, \forall j $, $\lambda_{1,l} \!\ge\!0, \forall l $ are the penalty parameters and function $h_l({\boldsymbol{x}_1}, \cdots ,{\boldsymbol{x}_n})= \sum\nolimits_{r \in [n]} \! \sum\nolimits_{k \in [d_r]} \! { {a_{r, k, l}^{(2)} x_{r,k}^2} \!+\! {b_{r, k, l}^{(2)} x_{r,k}} } - c_l^{(2)} - \varepsilon_2, \forall l $. \textcolor{black}{It is worth noting that while $\max\{h_l({\boldsymbol{x}_1},  \cdots  ,{\boldsymbol{x}_n}),0\}$ is non-differentiable at $h_l({\boldsymbol{x}_1},  \cdots  ,{\boldsymbol{x}_n})=0$, its squared version $[\max\{h_l({\boldsymbol{x}_1}, \cdots ,{\boldsymbol{x}_n}),0\}]^2$ is continuously differentiable. Its gradient is given by $\nabla [\max\{h_l({\boldsymbol{x}_1}, \!\cdots \!,{\boldsymbol{x}_n}),0\}]^2 = 2\max\{h_l({\boldsymbol{x}_1},\! \cdots \! ,{\boldsymbol{x}_n}),0\}\nabla h_l({\boldsymbol{x}_1}, \! \cdots \! ,{\boldsymbol{x}_n})$. Consequently, the penalty term in $q_{1,j}({\boldsymbol{x}_1},  \cdots  ,{\boldsymbol{x}_n})$ is inherently differentiable, which ensures that the overall function $q_{1,j}({\boldsymbol{x}_1},  \cdots  ,{\boldsymbol{x}_n})$ remains differentiable as long as $f_{1,j}({\boldsymbol{x}_1},  \cdots  ,{\boldsymbol{x}_n})$ is differentiable.}  We have demonstrated in Supplementary Material that the gap between the optimal objective values of the penalty-based unconstrained optimization problem in Eq. (\ref{eq:4_28_13}) and the original constrained optimization problem in Eq. (\ref{eq:result_L_cut_problem}) is bounded and will continuously decrease as $\lambda_{1,l}$ increase. \textcolor{black}{Since different workers may have conflicting objectives, a valid search direction should decrease all penalized local objectives simultaneously:}
\begin{equation}
\label{eq:12_19_6}
\begin{array}{l}
\!\! \mathop {\max }\limits_{j \in [M]} \left\{ {\frac{1}{\alpha }\! \left( {{q_{1,j}}({\boldsymbol{x}_1} \!-\! {\boldsymbol{s}_1}, \! \cdots \! ,{\boldsymbol{x}_n} \!-\! {\boldsymbol{s}_n}) - {q_{1,j}}({\boldsymbol{x}_1}, \! \cdots \! ,{\boldsymbol{x}_n})} \right)} \right\}    \! < \! 0, 
\end{array}
\end{equation}
where search direction $\boldsymbol{S}=[\boldsymbol{s}_1,\cdots,\boldsymbol{s}_n] \in \mathbb{R}^{\sum_i d_i} $ and $\boldsymbol{s}_i \in \mathbb{R}^{d_i} , \forall i \in [n]$. In order to find the valid search direction while protecting the data privacy, the developed distributed multi-gradient descent method proceeds in $(t+1)^{\rm{th}}$ iteration as follows.

\textbf{\emph{Local workers}} compute the gradients based on the information received from the master, i.e., $\nabla {q_{1,j}}(\boldsymbol{x}_1^t,\! \cdots \!,\boldsymbol{x}_n^t)$ in worker $j$, and transmit the gradients to the master. Upon obtaining the updated gradients from the workers, the \textbf{\emph{master}} proceeds with the following steps:

1. Addressing the following optimization problem:
\begin{equation}
\label{eq:12_8_eq6}
\begin{array}{l}
\mathop {\min }\limits_{\boldsymbol{p} \in \Delta } \left\langle {\sum\nolimits_{j \in [M]} {{p_j}\nabla {q_{1,j}^t}} ,\frac{1}{M}\sum\nolimits_{j \in [M]} {\nabla {q_{1,j}^t}} } \right\rangle 
\vspace{1mm}\\+ c||\frac{1}{M}\sum\nolimits_{j \in [M]} {\nabla {q_{1,j}^t}} || \cdot ||\sum\nolimits_{j \in [M]} {{p_j}\nabla {q_{1,j}^t}} ||
\end{array},
\end{equation}
where $\nabla q_{1,j}^t$ denotes the simplified form of gradient  $\nabla q_{1,j}(\boldsymbol{x}_1^t,\! \cdots \!,\boldsymbol{x}_n^t)$, $c>0$ is a constant, which will be discussed in detail in Appendix \ref{appendix:multi-gradient}. Variables $\boldsymbol{p}=[p_1,\cdots,p_M] \in \mathbb{R}^M$, and $\Delta=\{\boldsymbol{p}| 0 \le p_j\le 1, \forall j, \sum_{j} p_j = 1 \}$ is the simplex.  {\color{black}Problem (\ref{eq:12_8_eq6}) can be effectively addressed through utilizing interior point method, the arithmetic complexity of obtaining a $\gamma$-solution is $\mathcal{O}\left(\sqrt{2}((M+1)^3+(M+1)d_1^2)\log(\frac{1}{\gamma})\right)$ \cite{ben2021lectures}.}

2. After obtaining the optimal solution $\boldsymbol{p}^*$ of the problem in Eq. (\ref{eq:12_8_eq6}), the search direction and the variables can be respectively computed and updated as follows:
\begin{equation}
\label{eq:1_17_16}
\begin{array}{l}
{\boldsymbol{S}^{t + 1}} = \frac{{\sum\nolimits_{j \in [M]} {p_j^*\nabla {q_{1,j}^t}} }}{\phi^* } + \frac{1}{M}\sum\nolimits_{j \in [M]} {\nabla {q_{1,j}^t}} ,
\end{array}
\end{equation}
\begin{equation}
\label{eq:1_17_17}
\boldsymbol{x}_i^{t + 1} = \boldsymbol{x}_i^t - \alpha \boldsymbol{s}_i^{t + 1}, \forall i \in [n] ,
\end{equation}
where $\alpha$ is the step-size, $\phi^*  = \frac{1}{c}\frac{{||\sum\nolimits_{j} {{p_j^*}\nabla {q_{1,j}^t}} ||}}{{||\frac{1}{M}\sum\nolimits_{j} {\nabla {q_{1,j}^t}} ||}}$. The master then disseminates the updated variables $\boldsymbol{x}_i^{t + 1}, \forall i \in [n]$ to the workers. 
The detailed discussions of the distributed multi-gradient descent method are given in Appendix \ref{appendix:multi-gradient}.

\noindent \textcolor{black}{ \underline{\textbf{Discussion:}}
In this work, we aim to address the PLL problems with multiple objectives in a federated manner. Due to the composite nature of the proposed framework, we clarify which components are adapted from prior studies and which constitute our main novelties. Specifically, we follow the standard federated communication protocol to coordinate updates, adopt multi-gradient descent for handling multiple objectives, and utilize penalty-based reformulations motivated by classical constrained optimization. However, directly applying these existing ingredients to PLL is nontrivial, since PLL involves deeply nested hierarchical dependencies, and the federated multi-objective setting further introduces distributed computation and communication constraints. The main novelties of the proposed framework are summarized as follows.
\begin{enumerate}
   \item [*] \textbf{A Unified Federated PLL Framework:} As shown in Table~\ref{table:2}, existing methods usually address only part of the setting considered in this paper, such as trilevel optimization or multi-objective optimization. In contrast, the proposed FPLL provides a unified framework to simultaneously address poly-level nested structures, federated optimization, and multiple objectives.
    \item [*] \textbf{Level-wise $L$-cut Relaxation:} As highlighted in Table~\ref{table:cp}, the proposed $L$-cut provides a level-wise relaxation mechanism for approximating feasible regions defined by non-convex and non-linear hierarchical constraints, which are not directly handled by the traditional cutting plane methods. In addition, the level-wise design enables the proposed method to be recursively extended to higher-nested PLL problems, whereas most existing methods focus on bilevel or trilevel settings.
    \item [*] \textbf{Theoretical guarantees:} As demonstrated in Table~\ref{table:classic}, we provide non-asymptotic convergence guarantees for the proposed FPLL. Specifically, we establish the iteration complexity and communication complexity required to achieve an $\epsilon$-Pareto stationary point.
\end{enumerate} }

\section{Theoretical Analyses}

\begin{proposition}
\label{prop:1}
The feasible region of constraint ${g_i}({\boldsymbol{x}_1}, \cdots ,{\boldsymbol{x}_n}) = 0$ is a subset of $\boldsymbol{\mathcal{Q}}_i^{t+1}=\{\sum\nolimits_{r \in [n]}  \sum\nolimits_{k \in [d_r]}  { {a_{r, k, l}^{(i)} x_{r,k}^2} \!+\! {b_{r, k, l}^{(i)} x_{r,k}} } -c_l^{(i)} \le \varepsilon_i, l \! \in \! [|\boldsymbol{\mathcal{Q}}_{i}^{t+1}|] \}$. 
\end{proposition}

\vspace{2mm}

\textit{Proof:} See Appendix \ref{appendix:prop}.

\begin{proposition}
\label{prop:2} For any infeasible solution $({\boldsymbol{x}_1'}, \cdots ,{\boldsymbol{x}_n'})$ with respect to constraint ${g_i}({\boldsymbol{x}_1}, \cdots ,{\boldsymbol{x}_n}) \le \varepsilon_i$, the new generated $L$-cut can separate the infeasible solution from constraint ${g_i}({\boldsymbol{x}_1}, \cdots ,{\boldsymbol{x}_n})\le \varepsilon_i$, i.e., the new generated $L$-cut $\sum\nolimits_{r \in [n]} \! \sum\nolimits_{k \in [d_r]} \! { {a_{r, k, new}^{(i)} x_{r,k}^2} \!+\! {b_{r, k, new}^{(i)} x_{r,k}} } -c_{new}^{(i)}\! \le \! \varepsilon_i$ satisfies
\begin{equation}
  \left\{ \begin{array}{l}
 \sum\nolimits_{r \in [n]} \! \sum\nolimits_{k \in [d_r]} \! { {a_{r, k, new}^{(i)} x_{r,k}^2} \!+\! {b_{r, k, new}^{(i)} x_{r,k}} } \!-\! c_{new}^{(i)} \le  \varepsilon_i, \\ {\rm{if}}\,  {g_i}({\boldsymbol{x}_1}, \cdots{\boldsymbol{x}_n}) \le \varepsilon_i ,\\
\sum\nolimits_{r \in [n]} \! \sum\nolimits_{k \in [d_r]} \! { {a_{r, k, new}^{(i)} {x_{r,k}'^2}} \!+\! {b_{r, k, new}^{(i)} {x_{r,k}'}} }\! - \! c_{new}^{(i)} >  \varepsilon_i,
\end{array} \right.  
\end{equation}
where $\varepsilon_i \ge 0$ is a constant, thus, $\boldsymbol{\mathcal{Q}}_i^{t+1}$ converges monotonically as $i^{\rm{th}}$ level $L$-cuts are added.
\end{proposition}

\textit{Proof:} See Appendix \ref{appendix:prop}.

\begin{definition}
\label{def:1}
    \textbf{(Pareto Optimal Solution)} {\color{black} Following previous work \cite{fernando2022mitigating}, a solution $({\boldsymbol{x}_1^*}, \cdots ,{\boldsymbol{x}_n^*})$ is regarded as a Pareto optimal solution if and only if there is no feasible solution $({\boldsymbol{x}_1}, \cdots ,{\boldsymbol{x}_n})$ such that}
    \begin{equation}
    \begin{array}{l}
q_{1,j}({\boldsymbol{x}_1}, \cdots ,{\boldsymbol{x}_n}) \le q_{1,j}({\boldsymbol{x}_1^*}, \cdots ,{\boldsymbol{x}_n^*}), \forall j \in [M], \\
        \exists j, \; q_{1,j}({\boldsymbol{x}_1}, \cdots ,{\boldsymbol{x}_n}) < q_{1,j}({\boldsymbol{x}_1^*}, \cdots ,{\boldsymbol{x}_n^*}).
    \end{array}
    \end{equation}
\end{definition}
Finding a Pareto optimal solution when multi-objectives are non-convex is NP-hard \cite{yang2023federated,liu2021conflict}. Consequently, following \cite{liu2021conflict}, this work aims to find the Pareto stationary point as follows.

\begin{figure*}[t]
\centering
\subfigure[MNIST]
{\begin{minipage}{4.45cm}
\label{fig:MNIST_K1}
\includegraphics[scale=0.26]{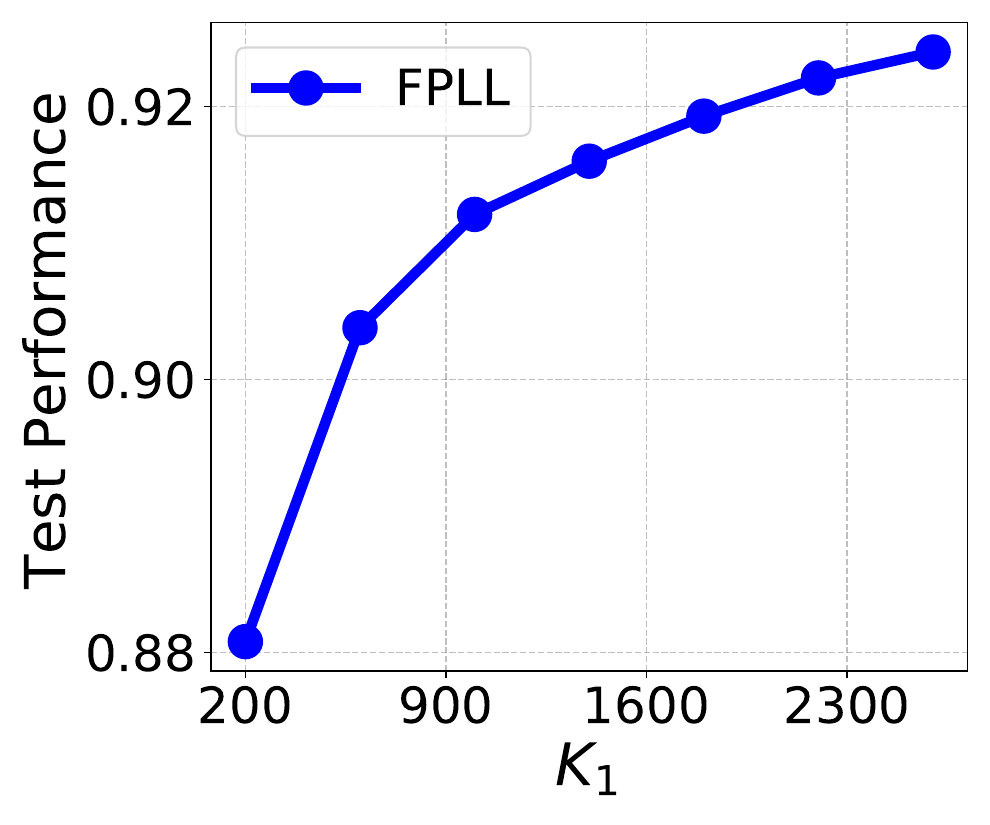}  
\end{minipage}}
\subfigure[KMNIST] 
{\begin{minipage}{4.45cm}
\label{fig:KMNIST_K1}   \includegraphics[scale=0.26]{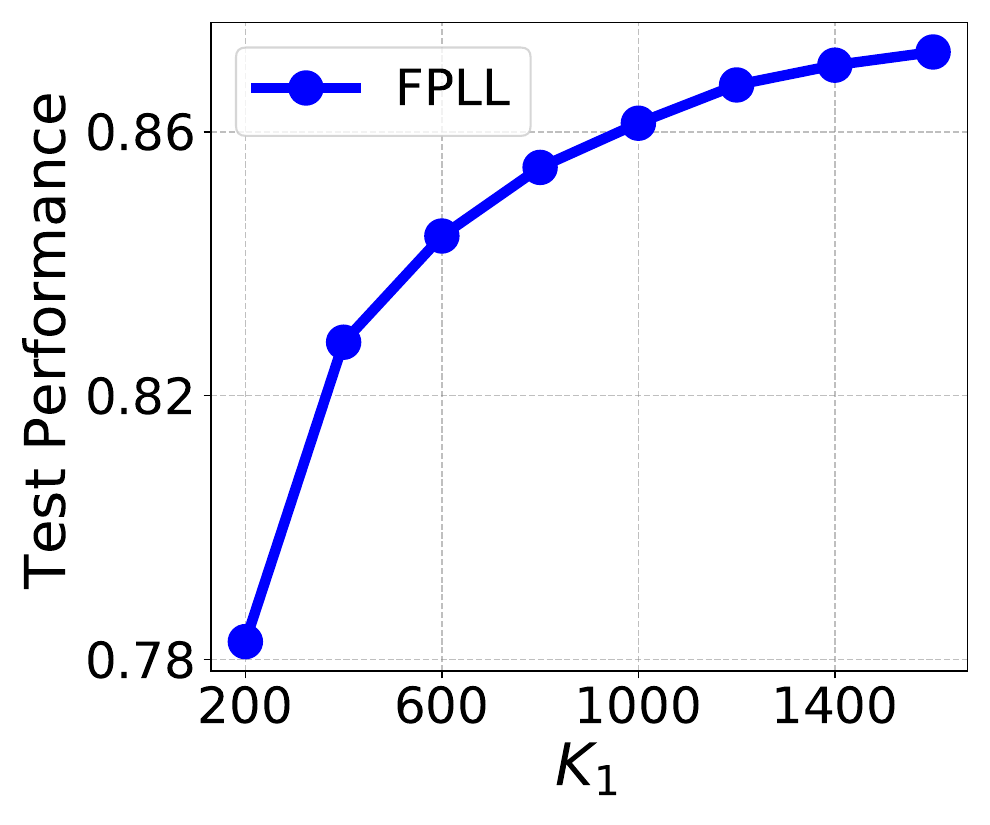}  
\end{minipage}}
\subfigure[QMNIST] 
{\begin{minipage}{4.45cm}
\label{fig:QMNIST_K1}
\includegraphics[scale=0.26]{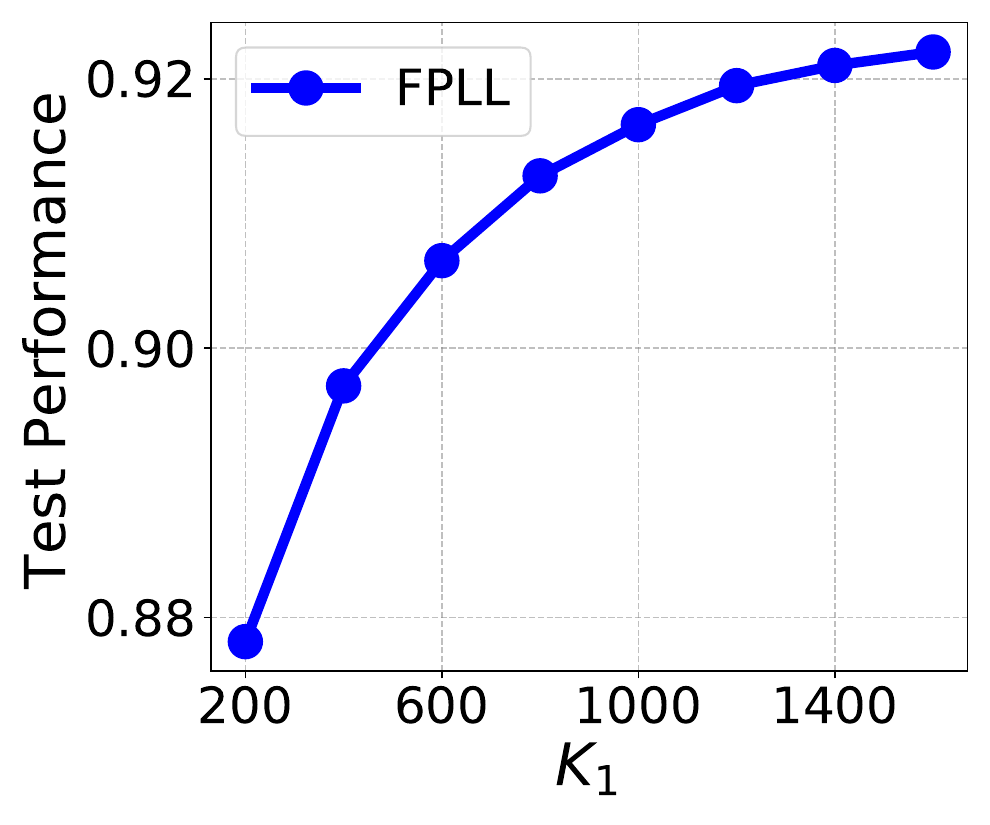}  
\end{minipage}}
\subfigure[F-MNIST] 
{\begin{minipage}{4.45cm}
\label{fig:Fashion_MNIST_K1}
\includegraphics[scale=0.26]{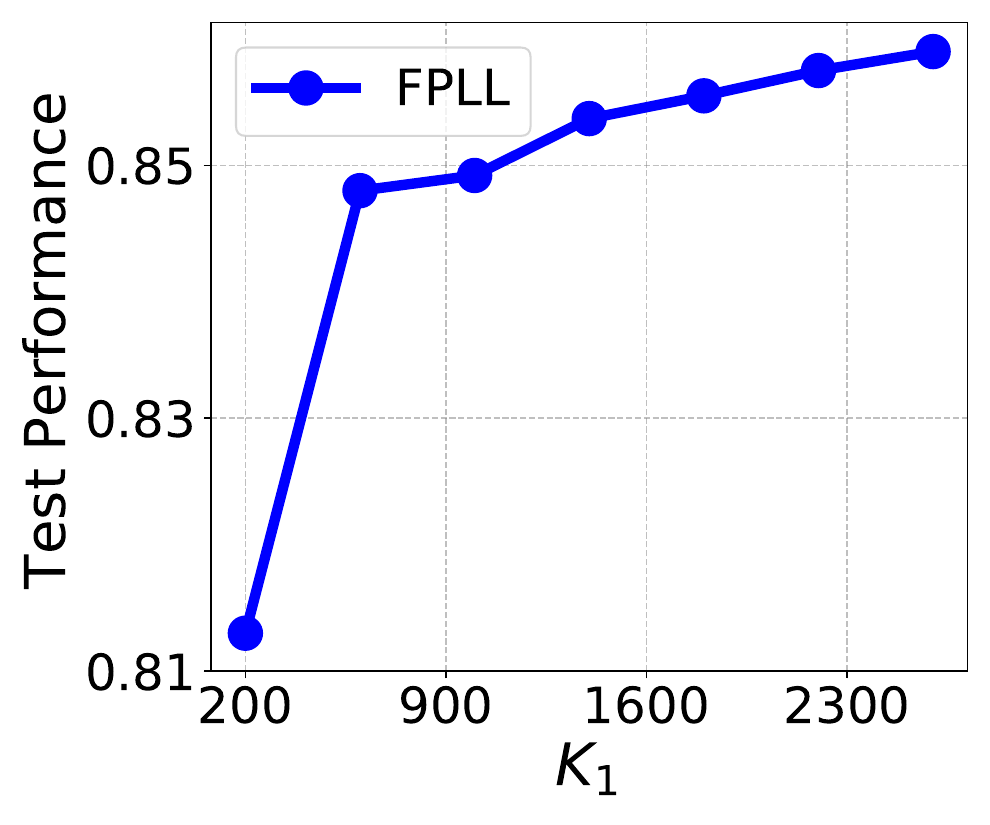}  
\end{minipage}}
\caption{The proposed FPLL allows for flexible control over the trade-off between performance and computational complexity through adjustments to $K_1$. The experiments are conducted on various datasets.} 
\label{fig:adjus_K1_1}
\end{figure*}

\begin{definition}
\label{def:2}
    \textbf{(Pareto Stationary Point)} Let $G^t=\mathop {\min }\nolimits_{\boldsymbol{p} \in \Delta } ||\sum\nolimits_{j \in [M]} {{p_j}\nabla {q_{1,j}}({\boldsymbol{x}_1^t}, \cdots ,{\boldsymbol{x}_n^t})} |{|^2}$. Following \cite{liu2021conflict,pan2023fedmdfg}, point $({\boldsymbol{x}_1^t}, \cdots ,{\boldsymbol{x}_n^t})$ is the Pareto stationary point if $ G^t = 0$, and it is called the $\epsilon$-Pareto stationary point if $G^t \le \epsilon$. Defining $T(\epsilon)$ as the initial iteration index where $G^t \le \epsilon$, i.e., $T(\epsilon)=\min\{ t | \,  G^t\le \epsilon\}$.
\end{definition}

\renewcommand\arraystretch{1.1}
\renewcommand\tabcolsep{11pt}
\begin{table*}[t]
\centering
\caption{Comparisons between the proposed FPLL and SOTA approaches in the continual learning task. Each experiment is repeated five times, with higher scores indicating better performance.}
{
\scalebox{1.09}{
\begin{tabular}{l|c|c|c|c|c}
\toprule
 Method   & MNIST    &  USPS & F-MNIST & KMNIST & QMNIST  \\ \hline

FGSM \cite{goodfellow2014explaining}   &  59.80 ± 1.43 $\%$  & 59.48 ± 1.31 $\%$   &  49.97 ± 1.55 $\%$   & 50.99 ± 0.27 $\%$  &  60.45 ± 0.37 $\%$ \\

PGD-7 \cite{madry2018towards}  &  60.51 ± 0.58 $\%$  & 59.79 ± 1.12 $\%$   &  51.04 ± 1.25 $\%$   & 50.48 ± 0.21 $\%$  & 59.73 ± 0.34 $\%$ \\

FMGDA \cite{yang2023federated}   & 54.05 ± 0.13 $\%$  & 55.49 ± 0.73 $\%$  & 40.06 ± 0.71 $\%$   &  42.81 ± 0.35 $\%$ & 55.98 ± 0.19 $\%$ \\

ADBO  \cite{jiao2022asynchronous}  &   56.37 ± 1.10 $\%$ & 59.01 ± 1.23 $\%$  &  47.91 ± 0.51 $\%$   & 48.70 ± 1.08 $\%$  &  57.05 ± 1.26 $\%$\\

FEDNEST \cite{tarzanagh2022fednest}    & 58.23 ± 2.60 $\%$  & 58.96 ± 0.74 $\%$  & 49.81 ± 1.85 $\%$   & 49.66 ± 0.79 $\%$   & 62.43 ± 0.34 $\%$  \\

FedAvg+EWC \cite{kirkpatrick2017overcoming}   &  67.18 ± 0.22 $\%$ &  73.04 ± 1.13 $\%$  & 50.87 ± 0.05 $\%$   & 53.42 ± 0.76 $\%$  & 66.91 ± 0.73 $\%$ \\

AFTO \cite{jiao2024provably}  &   77.85 ± 0.33 $\%$ &  78.62 ± 0.44 $\%$  & 69.63 ± 0.81 $\%$   & 69.31 ± 0.68 $\%$  & 77.90 ± 0.59 $\%$ \\

\hline
\textbf{FPLL}   & \textbf{92.97} ± \textbf{0.12} $\%$   & \textbf{94.95} ± \textbf{0.21} $\%$  &  \textbf{86.44} ± \textbf{0.17} $\%$ &  \textbf{87.30} ± \textbf{0.12} $\%$ & \textbf{92.41} ± \textbf{0.28} $\%$\\
\bottomrule  
\end{tabular}}
\label{tab:distributed-robust-CL}}
\end{table*}

\begin{assumption}
\label{assum:2}
 \textcolor{black}{ \textbf{(Boundedness)} Inspired by prior studies, e.g.,~\cite{lu2024first,hallak2026augmented,fu2023accelerated,ahn2020sgd,gurbuzbalaban2019convergence,davis2020stochastic}, we assume that the \textcolor{black}{iterates $\{\boldsymbol{x}_1^t,\boldsymbol{x}_2^t,\cdots,\boldsymbol{x}_n^t\}_{t\ge1}$ are bounded,} i.e., $||\boldsymbol{x}_i^t||^2 \le \kappa_i,i \in [n]$.} Furthermore, we assume that the objective function is bounded from below, i.e., $q_{1,j}^*=\inf {q_{1,j}}({\boldsymbol{x}_1}, \cdots ,{\boldsymbol{x}_n}) > -\infty$.
\end{assumption}


\underline{\textit{Discussion.}} \textcolor{black}{It is worth noting that Assumption 1 is introduced primarily to facilitate the theoretical convergence analysis, whereas the practical implementation of the proposed method does not strictly depend on the fulfillment of this assumption.} It is seen from Eq. (\ref{eq:4_28_13}) that term $\sum\nolimits_{l=1}^{|\boldsymbol{\mathcal{Q}}_2^{t+1}|} \lambda_{1,l} [\max\{\sum\limits_{r \in [n]} \! \sum\limits_{k \in [d_r]} \! { {a_{r, k, l}^{(2)} x_{r,k}^2} \!+\! {b_{r, k, l}^{(2)} x_{r,k}} }- c_l^{(2)} -\varepsilon_2, 0\}]^2 \ge 0$ since penalty parameters $\lambda_{1,l} \ge 0$. Thus, we can obtain that $q_{1,j}^* > -\infty, \forall j$ if $f_{1,j}^* > -\infty, \forall j$, which is a mild and commonly employed assumption \cite{liu2021conflict}.

\begin{assumption}
\label{assum:1}
   \textbf{(Smoothness)} Following \cite{ji2021bilevel,yang2021provably}, we assume that ${q_{1,j}}, \forall j \in [M]$ have $L$-Lipschitz continuous gradient, that is, for any $({\boldsymbol{x}_1}, \cdots ,{\boldsymbol{x}_n})$ and $({\hat{\boldsymbol{x}}_1}, \cdots ,{\hat{\boldsymbol{x}}_n})$, it holds that
   \begin{equation}
   \begin{array}{l}
      || \nabla {q_{1,j}}({\boldsymbol{x}_1}, \cdots ,{\boldsymbol{x}_n}) - \nabla {q_{1,j}}({\hat{\boldsymbol{x}}_1}, \cdots ,{\hat{\boldsymbol{x}}_n})||  \\ \le L || [\boldsymbol{x}_1 - \hat{\boldsymbol{x}}_1;\cdots;\boldsymbol{x}_n - \hat{\boldsymbol{x}}_n ]||, \; \forall j.
   \end{array}
   \end{equation}
\end{assumption}

\underline{\textit{Discussion.}} In level-wise $L$-cut relaxation, the continuous updating of $L$-cut throughout the iterative process often renders it impractical for real-world application scenarios. The continuous addition of $L$-cut may lead to memory overhead, a situation also observed in the field of continual learning \cite{yoon2021federated}. To address this concern and achieve adaptable control over the level-wise $L$-cut relaxation, we stipulate that the level-wise $L$-cut relaxation remains fixed after $K_1$ iterations, where $K_1>0$ is a constant. Thus, there are at most $\lfloor \frac{K_1}{\tau} \rfloor$ $2^{\rm{nd}}$ layer $L$-cuts. By combining that all variables are bounded, we have that $\sum\nolimits_{l=1}^{|\boldsymbol{\mathcal{Q}}_2^{t+1}|} \lambda_{1,l} [\max\{\sum\limits_{r \in [n]} \! \sum\limits_{k \in [d_r]} \! { {a_{r, k, l}^{(2)} x_{r,k}^2} \!+\! {b_{r, k, l}^{(2)} x_{r,k}} }- c_l^{(2)} -\varepsilon_2, 0\}]^2$ satisfies the gradient Lipschitz condition. 
As long as the first-level objectives ${{f_{1,j}}({\boldsymbol{x}_1}, \cdots ,{\boldsymbol{x}_n})}$ adhere to the gradient Lipschitz condition, which is a commonly used and mild assumption in machine learning \cite{ji2021bilevel,yang2021provably}, ${{q_{1,j}}({\boldsymbol{x}_1}, \cdots ,{\boldsymbol{x}_n})}$ will consistently satisfy the gradient Lipschitz condition.

\begin{theorem}
\label{theorem:1}
    \textbf{(Iteration Complexity)} Let constant step-size $\alpha \le \frac{1}{L}$ and constant  $0<c<1$, under Assumptions \ref{assum:2}  and \ref{assum:1}, the upper bound for the number of iterations for the proposed FPLL to obtain the $\epsilon$-Pareto stationary point is given by
\begin{equation}
\label{eq:theorem}
T(\epsilon)\sim  \frac{ \frac{1}{M}\sum\limits_{j \in [M]} \max\limits_{k \in [K_1]} {{q_{1,j}}(x_1^{k}, \! \cdots \! ,x_n^{k})}  -  q_{1,j}^*  }{{\frac{\alpha }{2}\left( {1 - {c^2}} \right) \epsilon}} + K_1 - 1 ,
\end{equation}
where $K_1 > 0$ is a constant that regulates the trade-off between the complexity and performance, $M>0$ is the number of workers in distributed systems, $q_{1,j}^*$ is the lower-bound of $q_{1,j}$ as discussed in Assumption \ref{assum:2}.

\textit{Proof:} See Appendix \ref{appendix:proof-theorem}.

\underline{\textit{Discussion.}} It can be seen from Eq. (\ref{eq:theorem}) that the iteration complexity depends on the choice of $K_1$. Specifically, a larger $K_1$ results in a larger term $K_1-1$ in Eq. (\ref{eq:theorem}) as well as a larger value of $\max\limits_{k \in [K_1]} {{q_{1,j}}(x_1^{k}, \cdots ,x_n^{k})}$, thereby leading to a higher iteration complexity for achieving an $\epsilon$-Pareto stationary point. Additionally, it can be seen from Propositions \ref{prop:1} and \ref{prop:2} that we can obtain a better level-wise relaxation as $L$-cuts are added. Thus, the constant $K_1$ also governs the approximation capability of the proposed FPLL, indicating that a larger $K_1$ results in a more accurate approximation. \textcolor{black}{In addition, $\tau$ affects the practical convergence speed and the tightness of the resulting relaxation by controlling the frequency of $L$-cut updates. However, our iteration-complexity analysis focuses on the worst-case scenario: it treats the first $K_1$ iterations as a refinement phase, without explicitly accounting for the number or frequency of $L$-cut updates. Therefore, $K_1$, rather than $\tau$, appears explicitly in the worst-case iteration-complexity bound.} It is important to mention that the value of $K_1$ can be adjusted flexibly in the proposed framework FPLL to align with the computational and communication capacities of the distributed system. Consequently, the proposed framework FPLL offers flexibility in adjusting the trade-off between computational complexity and performance.
\end{theorem}

{\color{black}\begin{theorem}
\label{theorem:2}
\textbf{(Communication Complexity)} According to prior work \cite{backes2013asynchronous}, the communication complexity is defined as the total number of bits transmitted during the communication process. In this work, the communication complexity is composed of two primary parts: (i) the cost of transmitting gradients and model parameters at each iteration, and (ii) the communication overhead associated with updating the $L$-cuts. Consequently, the communication complexity of the proposed method is
\begin{equation}
\begin{array}{l}
32(2\sum\nolimits_{i=1}^nd_i)MT(\epsilon)\!+\!32(3\sum\nolimits_{i=1}^nd_i+1)nM\lfloor \frac{K_1}{\tau} \rfloor.
\end{array}
\end{equation}

\textit{Proof:} See Appendix \ref{appendix:proof-theorem2}.
\end{theorem}}

\textcolor{black}{\underline{\textit{Discussion.}}  In contrast to the iteration complexity, the communication complexity explicitly depends on $\tau$, since each $L$-cut update incurs additional communication and the $L$-cuts are updated at most $\lfloor K_1/\tau\rfloor$ times.}

\begin{figure*}[t]
\centering
\subfigure[MNIST $\rightarrow$ F-MNIST]
{\begin{minipage}{8.2cm}
\label{fig:FT_1}
\includegraphics[scale=0.23]{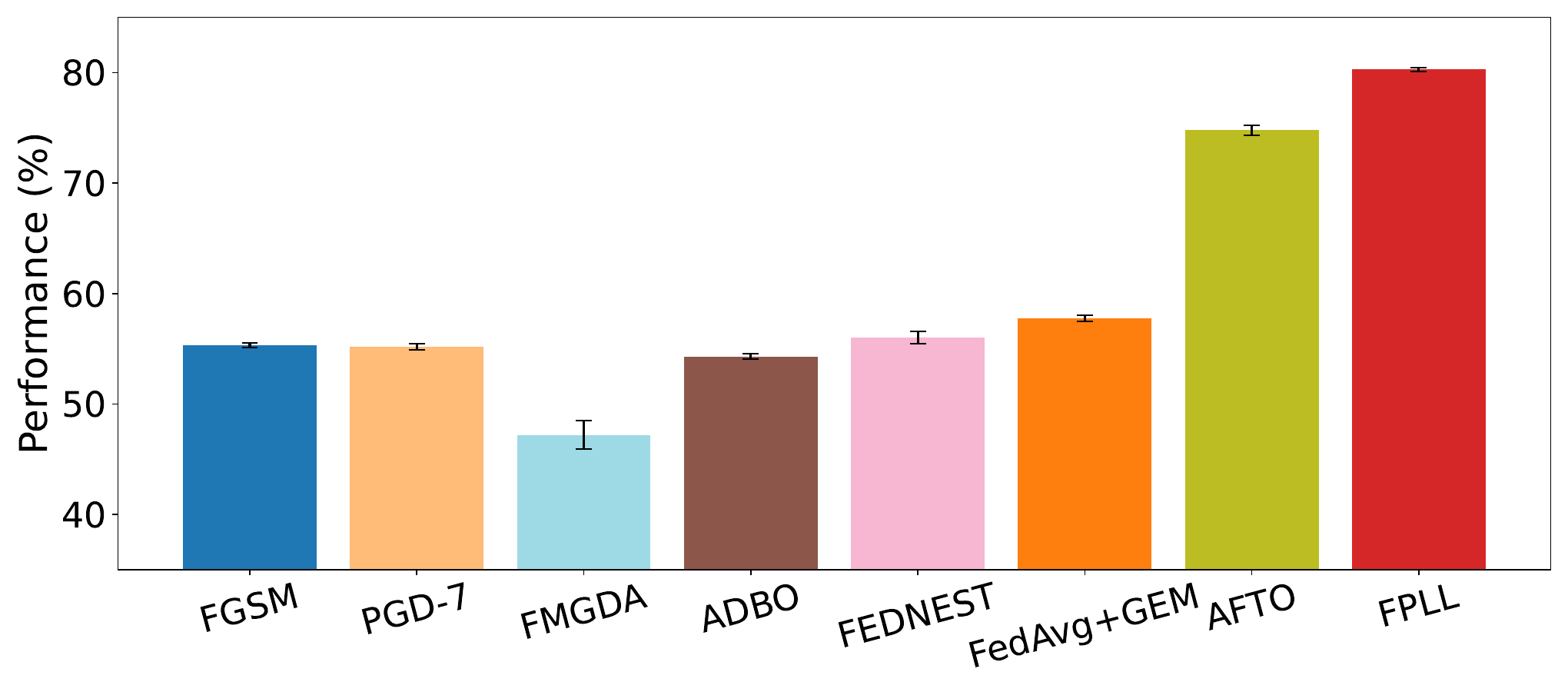}  
\end{minipage}}
\subfigure[MNIST $\rightarrow$ USPS] 
{\begin{minipage}{8.2cm}
\label{fig:FT_2}   \includegraphics[scale=0.23]{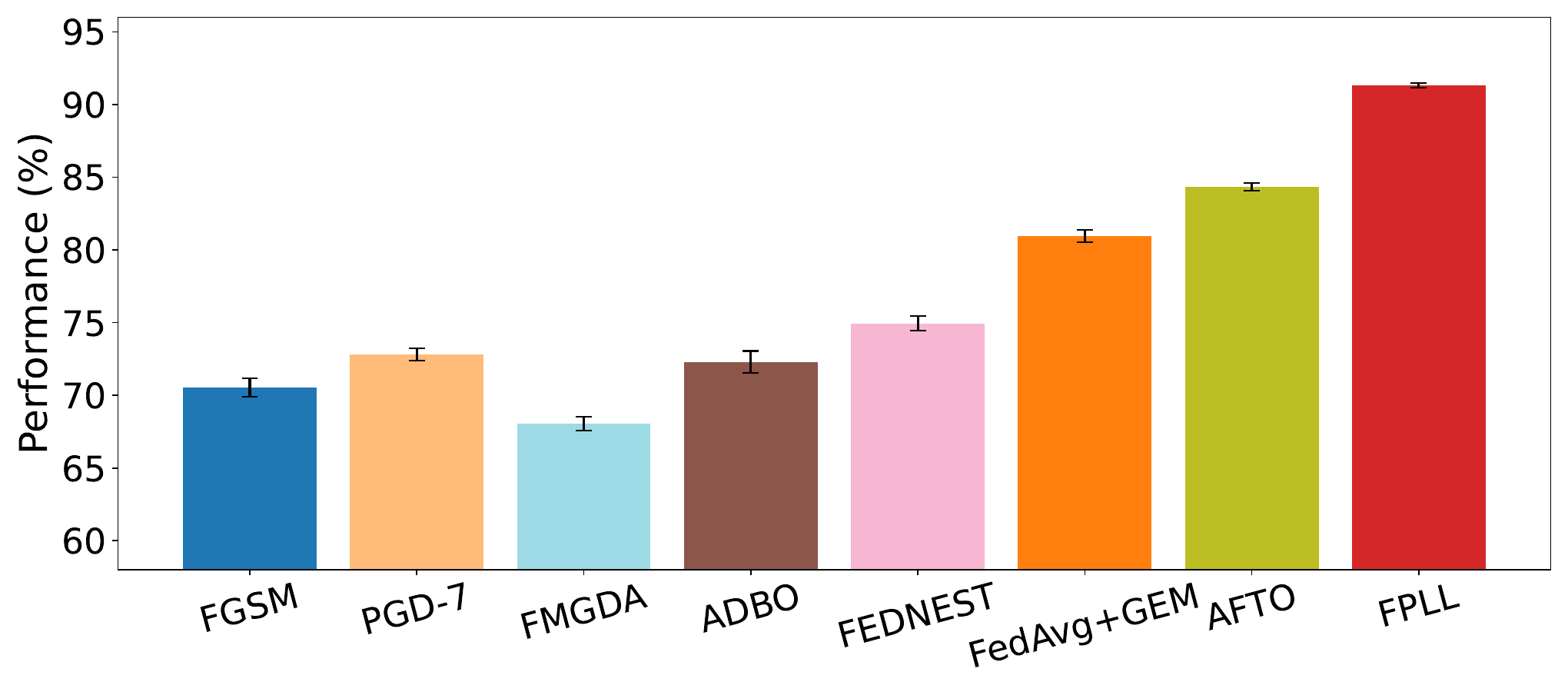}  
\end{minipage}}
\subfigure[F-MNIST $\rightarrow$ USPS] 
{\begin{minipage}{8.2cm}
\label{fig:FT_3}
\includegraphics[scale=0.23]{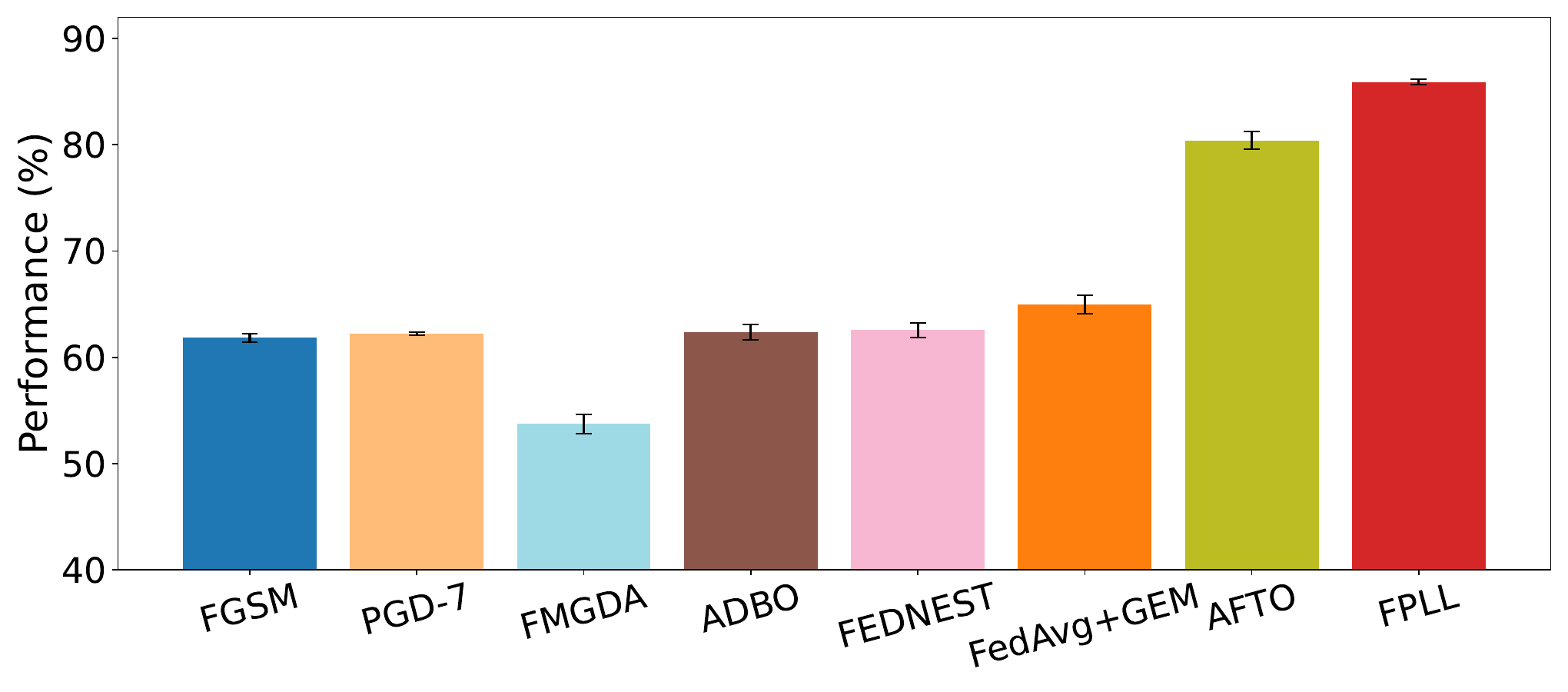}  
\end{minipage}}
\subfigure[USPS $\rightarrow$ MNIST ] 
{\begin{minipage}{8.2cm}
\label{fig:FT_4}
\includegraphics[scale=0.23]{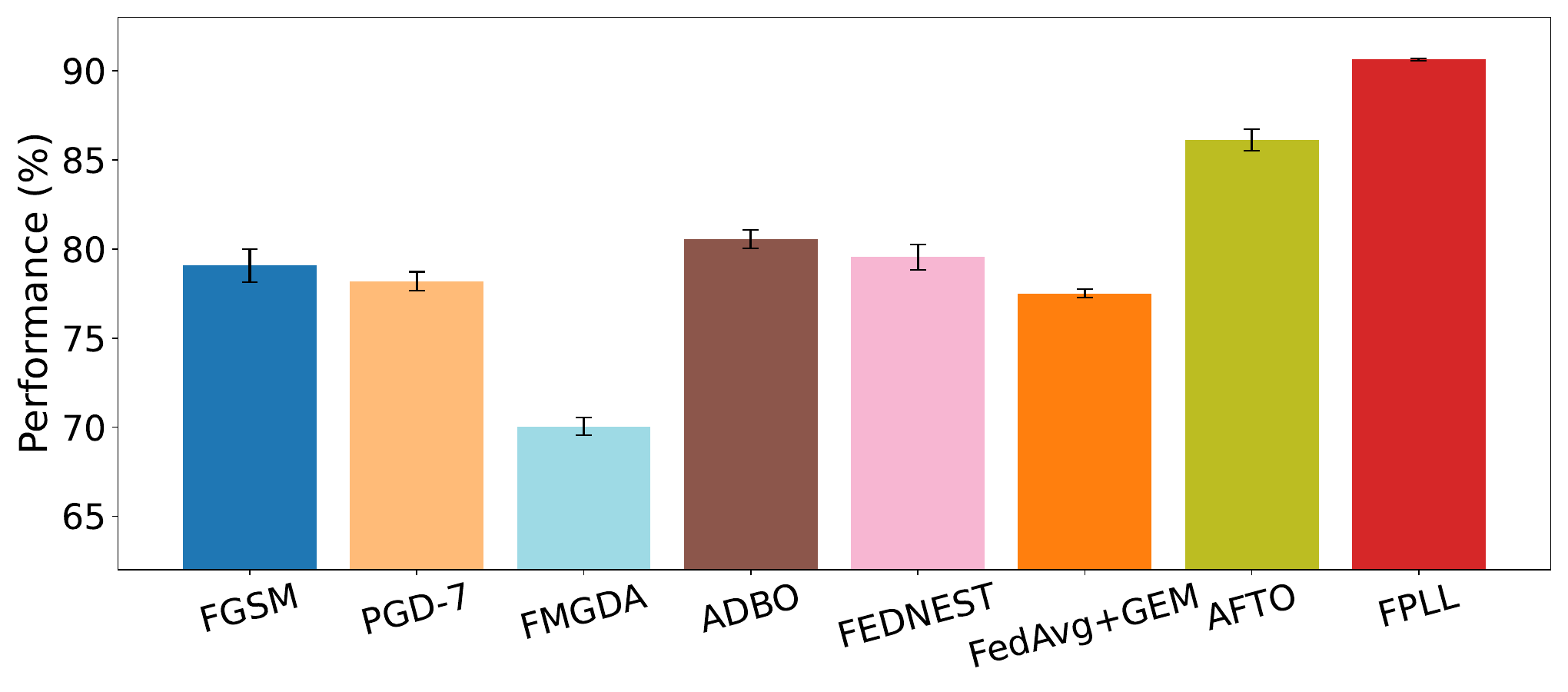}  
\end{minipage}}
\caption{Performance comparisons between the proposed FPLL and the SOTA approaches in the fine-tuning task. A $\rightarrow$ B represents that dataset A is used to pretrain the models and dataset B is used to fine-tune.} 
\end{figure*}

\section{Experiments}
\label{sec:experiments}
In the experiment, the distributed fine-tuning \cite{tian2023trainable} and distributed continual learning \cite{peng2023ideal} tasks are implemented to assess the performance of the proposed framework. The proposed FPLL is compared with the state-of-the-art federated bilevel optimization approaches FEDNEST \cite{tarzanagh2022fednest}, ADBO \cite{jiao2022asynchronous},  federated trilevel optimization method AFTO \cite{jiao2024provably}, and federated multi-objective optimization method FMGDA \cite{yang2023federated}. The adversarial learning methods FGSM \cite{goodfellow2014explaining} and PGD-7 \cite{madry2018towards} are used as the baseline methods. Specifically, we assess the effectiveness of the proposed FPLL to tackle the poly-level optimization problems in a federated manner and examine its flexibility in controlling the trade-off between computational complexity and performance. \textcolor{black}{To ensure a fair comparison, we independently tune the learning rate for each method based on its validation performance, searching over $\{0.0005, 0.001, 0.0025, 0.005, 0.01, 0.025, 0.05\}$. All methods use the same fixed validation set and evaluation protocol, with validation performance evaluated every 50 communication rounds. For the continual learning task, the learning rate that maximizes the average clean and PGD-7 accuracy across the previous and new tasks is selected. For the fine-tuning task, the learning rate that maximizes the average of the source-domain clean accuracy, target-domain clean accuracy, and target-domain PGD-7 accuracy is selected. The number of communication rounds is fixed to be the same across all methods, ensuring that all methods are evaluated under a common communication-round budget. Each experiment is repeated five times, and the results are reported as the mean $\pm$ standard deviation over the five runs. Detailed experimental settings are provided in the Supplementary Material: Tables S2 and S4 report the tuned method-specific learning rates for the continual learning and fine-tuning tasks, respectively. Tables S3 and S5 summarize the corresponding shared experimental configurations, including the number of communication rounds, the number of workers, and the adversarial attack settings. Table S6 and S7 report the method-specific computational costs, including local and total gradient evaluations, and runtime.} All models are implemented in PyTorch, with experiments performed on a server featuring two NVIDIA 4090 GPUs.

\noindent\textbf{Baseline Methods.} 

\begin{enumerate}

    \item [*] FEDNEST \cite{tarzanagh2022fednest} is a state-of-the-art federated nested optimization method. FEDNEST extends standard federated learning to tackle bilevel programming, employing federated alternating stochastic gradients and federated hypergradient computation.

    \item [*] ADBO \cite{jiao2022asynchronous} is a state-of-the-art distributed bilevel optimization method. In ADBO, the lower-level optimization problem is regarded as a constraint to the upper-level optimization problem, and the cutting plane method based on convex optimization is used to approximate the lower-level constraint.

    \item [*] AFTO \cite{jiao2024provably} is a state-of-the-art distributed trilevel optimization framework. In AFTO, the hyper-polyhedral approximation based on $\mu$-weakly convex functions is proposed to approximate the original feasible region in trilevel optimization problems.

    \item [*] FMGDA \cite{yang2023federated} is a state-of-the-art federated multi-objective single-level optimization approach, which can be utilized to address the multi-objective optimization problems while keeping data privacy.

    \item [*] FGSM \cite{goodfellow2014explaining} is a method used to generate adversarial examples by slightly shifting the input data along the gradient of the loss function, using a small scaling factor for the perturbation, making it fast and efficient.

    \item [*] PGD \cite{madry2018towards} is a more advanced attack that iteratively applies small perturbations in the direction of the gradient, while projecting the perturbed input back into a valid region. This method is more robust and can generate stronger adversarial examples compared with FGSM.

    \item [*] \textcolor{black}{FedAvg combined with Elastic Weight Consolidation (EWC) \cite{kirkpatrick2017overcoming} is used as a baseline method, termed FedAvg+EWC, for distributed continual learning. For distributed fine-tuning, to preserve the performance on the pretraining task, FedAvg is combined with Gradient Episodic Memory (GEM) \cite{guirguis2022cfa} as another baseline method, termed FedAvg+GEM.}

\end{enumerate}

\begin{figure}[t]
\centering
\subfigure[Test Performance]
{\begin{minipage}{4.3cm}
\includegraphics[scale=0.26]{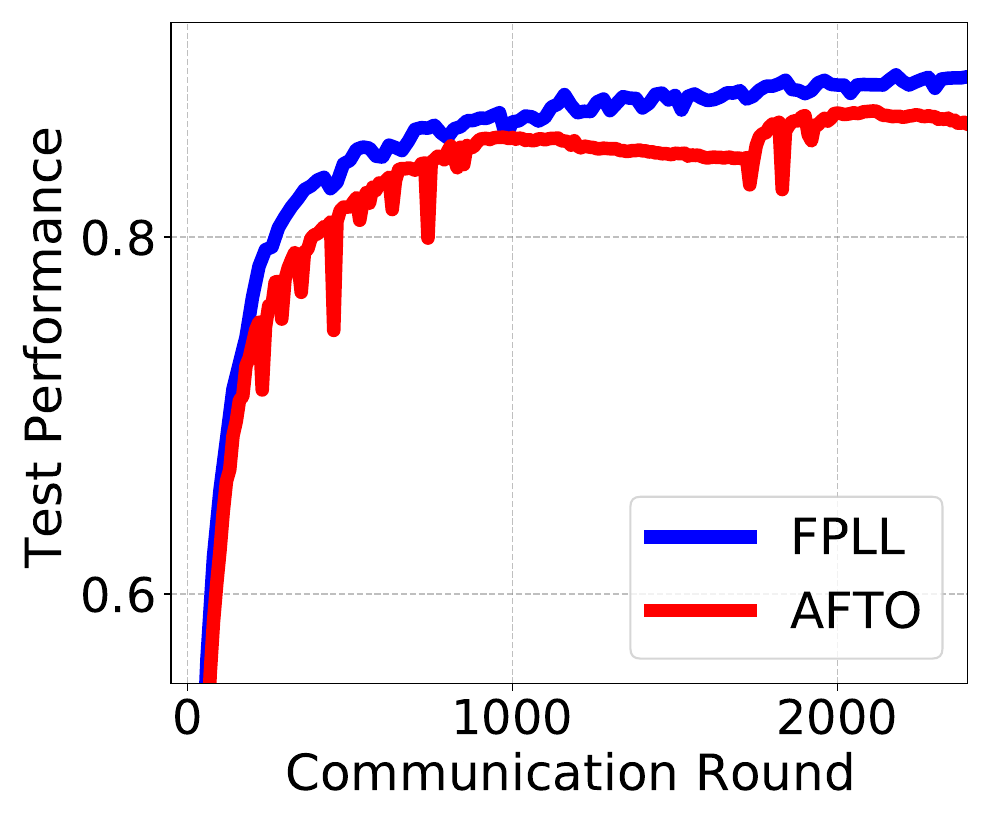}  
\end{minipage}}
\subfigure[Test Loss] 
{\begin{minipage}{4.3cm} \includegraphics[scale=0.26]{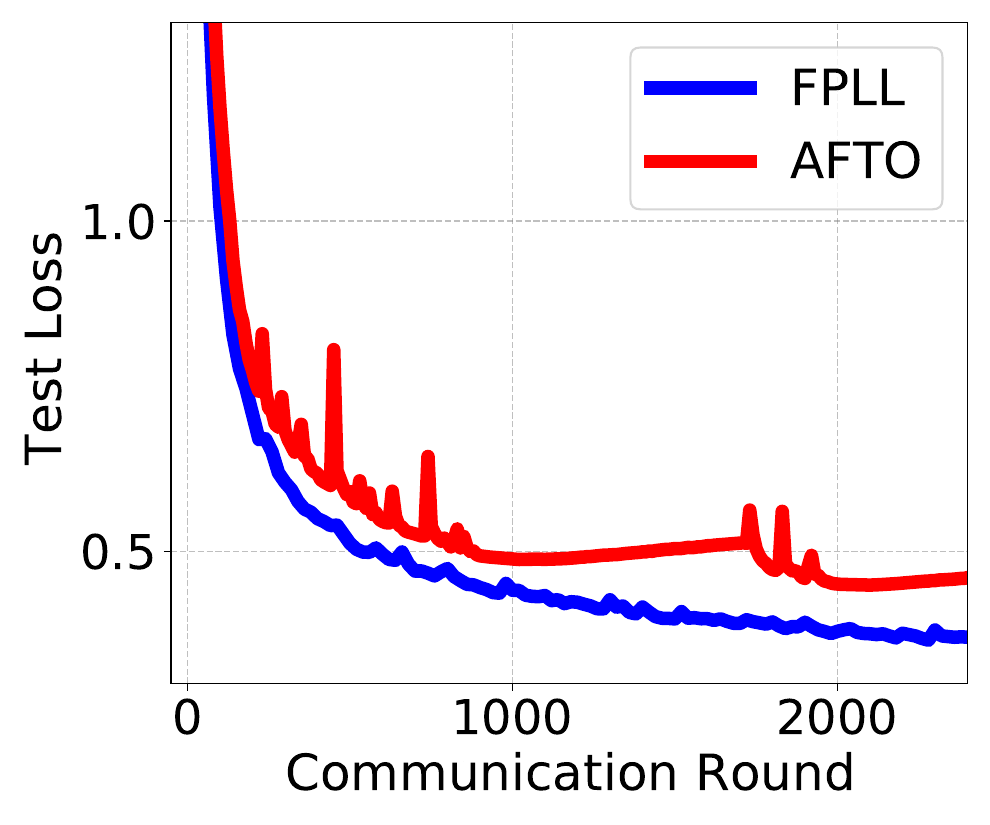}  
\end{minipage}}
\caption{{\color{black}Comparison between the proposed FPLL and the state-of-the-art distributed trilevel optimization method AFTO on the USPS $\rightarrow$ MNIST task in terms of the number of communication rounds.}}
\label{fig:compare_afto}
\end{figure}

\renewcommand\arraystretch{1.1}
\renewcommand\tabcolsep{11pt}
\begin{table}[t]
\centering
\caption{Comparisons between FPLL with $L$-cuts (FPLL) against FPLL with $\mu$-cuts (FPLL-). Experiments are conducted five times, with higher scores indicating superior performance.}
{
\scalebox{1.09}{
\begin{tabular}{l|c|c}
\toprule
 Datasets   & FPLL-    &  FPLL   \\ \hline

MNIST  & 87.23 ± 0.34 $\%$ &  \textbf{92.97} ± \textbf{0.12} $\%$ \\

USPS  &  90.11 ± 0.27 $\%$ & \textbf{94.95} ± \textbf{0.21} $\%$  \\

F-MNIST  & 82.07 ± 0.31 $\%$  &  \textbf{86.44} ± \textbf{0.17} $\%$ \\

KMNIST  &  83.61 ± 0.29 $\%$ & \textbf{87.30} ± \textbf{0.12} $\%$  \\

QMNIST  &  87.93 ± 0.24 $\%$ & \textbf{92.41} ± \textbf{0.28} $\%$  \\

\bottomrule  
\end{tabular}}
\label{tab:compare_mu}}
\end{table}

\subsection{Continual Learning}

Continual Learning explores strategies for learning on a sequence of tasks without encountering catastrophic forgetting \cite{rusu2016progressive,yoon2021federated} of previously acquired knowledge, i.e., previous tasks, which can be expressed as a bilevel optimization problem \cite{peng2023ideal}. To improve the robustness of the machine learning model when conducting continual learning, the adversarial architecture-based continual learning task is considered inspired by \cite{rusu2016progressive,ebrahimi2020adversarial}. The distributed adversarial continual learning problem can be formulated as
\begin{equation}
\! \begin{array}{l}
\! \min  \! {\left( \! { \sum\limits_{(x,y)  \in  D_1^{\rm{B}}} \! \frac{{{\cal L}([{\boldsymbol{\theta} _1},{\boldsymbol{\theta} _2}],x,y, {\boldsymbol{P}_2})}}{|D_1^{\rm{B}}|} ,\!  \cdots \!  ,\! \sum\limits_{(x,y) \in D_M^{\rm{B}}} \! \frac{{{\cal L}([{\boldsymbol{\theta} _1},{\boldsymbol{\theta} _2}],x,y, {\boldsymbol{P}_2})}}{|D_M^{\rm{B}}|}  } \! \right)^{\top}}\\

\! {\rm{  s}}{\rm{.t}}{\rm{.}}  {\boldsymbol{P}_2} \!=\! \mathop {\arg \max }\limits_{{\boldsymbol{P}_2'}} \sum\limits_{j=1}^M  \sum\limits_{(x,y) \in D_j^{\rm{B}}} \frac{{\cal L}([{\boldsymbol{\theta} _1},{\boldsymbol{\theta} _2}],x,y, {\boldsymbol{P}_2'})}{{|{D_j^{\rm{B}}}|}} + \mathcal{K} (\boldsymbol{P}_2')\\

\quad {\rm{  s}}{\rm{.t}}{\rm{.}} \; {\boldsymbol{\theta}_1} = \mathop {\arg \min }\limits_{{\boldsymbol{\theta}_1'}} \sum\limits_{j=1}^M
\sum\limits_{(x,y) \in D_j^{\rm{A}}} \frac{{\cal L}({\boldsymbol{\theta}_1'},x,y, {\boldsymbol{P}_1})}{{|{D_j^{\rm{A}}}|}}\\

\qquad  {\rm{  s}}{\rm{.t}}{\rm{.}} \; {\boldsymbol{P}_1} = \mathop {\arg \max }\limits_{{\boldsymbol{P}_1'}} \sum\limits_{j=1}^M \sum\limits_{(x,y) \in D_j^{\rm{A}}}  \frac{{\cal L}({\boldsymbol{\theta}_1'},x,y, {\boldsymbol{P}_1'})}{{|{D_j^{\rm{A}}}|}} + \mathcal{K} (\boldsymbol{P}_1')
\\

{\mathop{\rm var}} .\qquad \qquad \qquad {\boldsymbol{\theta} _1},{\boldsymbol{\theta} _2},{\boldsymbol{P}_1},{\boldsymbol{P}_2},
\end{array}
\end{equation}
where ${\boldsymbol{\theta} _1},{\boldsymbol{\theta} _2}$ represent the model parameters corresponding to previous task A and new task B, ${\boldsymbol{P}_1},{\boldsymbol{P}_2}$ represent the adversarial noise in tasks A, B. $ {D_j^{\rm{A}}},  {D_j^{\rm{B}}} $ denote the datasets in task A, B on $j^{\rm{th}}$ worker. $\mathcal{K} (\boldsymbol{P}_2)
= \sum_{k}  r_2[\max\{ -p_{2,k}-\varepsilon, p_{2,k}- \varepsilon, 0 \}]^2$, $\mathcal{K} (\boldsymbol{P}_1)
= \sum_{k}  r_1[\max\{ -p_{1,k}-\varepsilon, p_{1,k}- \varepsilon, 0 \}]^2$ are the penalty terms for the $L_{\infty}$-ball constraints, where $r_1>0, r_2>0$ are constants. In the experiments, following \cite{abati2020conditional}, the efficacy of the proposed FPLL is assessed through the continual learning on diverse digit recognition datasets, that is, MNIST, USPS, Fashion MNIST, KMNIST, and QMNIST. Since we aim to mitigate the catastrophic forgetting and improve the robustness of machine learning models, the average performance across the accuracy and robustness of the method on the previous task A and new task B is used as the metric. In the experiments, we set $M=20$, and we utilize the PGD-7 attack \cite{madry2018towards} to evaluate the robustness. Each dataset is partitioned into two subsets, each consisting of consecutive classes. This division gives rise to two distinct five-class classification tasks, which are sequentially observed as the previous task and the new task. The comparisons between the proposed FPLL and the SOTA (State-Of-The-Art) approaches are reported in Table \ref{tab:distributed-robust-CL}. It is seen from Table \ref{tab:distributed-robust-CL} that the performance of the proposed FPLL with $L$-cuts surpasses that of the SOTA approaches mainly because: 1) FPLL is capable of effectively tackling the poly-level learning problems in a federated manner while the baseline approaches can only be applied to the distributed bilevel and trilevel optimization problems; 2) the utilization of $L$-cuts and distributed multi-gradient descent method facilitates the resolution of poly-level learning problems with multiple objectives.

To further assess the performance of the proposed $L$-cuts, FPLL with $L$-cuts is compared against FPLL with $\mu$-cuts \cite{jiao2024provably}. As demonstrated in Table \ref{tab:compare_mu}, FPLL with $L$-cuts outperforms FPLL- (i.e., FPLL with $\mu$-cuts), highlighting the superiority of the proposed $L$-cuts.

Within the proposed FPLL, we can flexibly control the trade-off between the computational complexity and performance by adjusting $K_1$. Specifically, if communication and computational capabilities are limited in a distributed system, a relatively small $K_1$ can be set, necessitating fewer iterations to converge, as discussed in Theorem \ref{theorem:1}. Conversely, if superior performance is a priority, a larger $K_1$ can be configured to construct a more refined level-wise $L$-cut relaxation. As depicted in Fig. \ref{fig:adjus_K1_1}, the performance of FPLL steadily enhances with an increase in $K_1$.

\subsection{Fine-tuning
}

Fine-tuning machine learning models for other tasks typically yields inferior out-of-distribution generalization, as the model tends to overfit to the new data, thereby risking the loss of previously acquired pre-trained features \cite{tian2023trainable}. Inspired by the bilevel fine-tuning in \cite{tian2023trainable} and adversarial learning \cite{ebrahimi2020adversarial}, the robust fine-tuning task, which aims to simultaneously maintain the generalization capability acquired during the pretraining of the model while enhancing the performance and adversarial robustness of the model on the new task, is considered.
The federated robust fine-tuning can be formulated as
\begin{equation}
\label{eq:1_9_21}
\begin{array}{l}
\mathop {\min } {\left( {\sum\limits_{(x,y) \in D_1^{\rm{val}}} \frac{{{\cal L}(x,y,\boldsymbol{\theta} )}}{|D_1^{\rm{val}}|}  , \cdots ,\sum\limits_{(x,y) \in D_M^{\rm{val}}} \frac{{{\cal L}(x,y,\boldsymbol{\theta} )}}{|D_M^{\rm{val}}|}  } \right)^{\top}}\\

{\rm{s}}{\rm{.t}}{\rm{.  }} \quad \boldsymbol{\theta} = \mathop {\arg \min }\limits_{\boldsymbol{\theta'} } \sum\limits_{j = 1}^M 
{\sum\limits_{(x,y) \in D_j^{\rm{tr}}} \! \frac{{{\cal L}(x,y,\boldsymbol{\theta'} ,\boldsymbol{P})}}{{|D_j^{\rm{tr}}|}}  \!+\! \gamma {\rm{||}}\boldsymbol{\theta'}  - {\boldsymbol{\theta}^{\rm{pre}}}|{|^2}} \\

\quad{\rm{s}}{\rm{.t}}{\rm{.  }} \quad \boldsymbol{P} = \mathop {\arg \max }\limits_{\boldsymbol{P}'} \sum\limits_{j = 1}^M {\sum\limits_{(x,y) \in D_j^{\rm{tr}}} \! \frac{{{\cal L}(x,y,\boldsymbol{\theta}' ,\boldsymbol{P}')}}{{|D_j^{\rm{tr}}|}} \!+\! \mathcal{Q}(\boldsymbol{P}')}
 \\
{\mathop{\rm var}} . \qquad \qquad \qquad \gamma ,\boldsymbol{\theta} ,\boldsymbol{P},
\end{array}
\end{equation}
where $\boldsymbol{\theta}$, $\boldsymbol{P}$, and $\gamma$ denote the model parameters, adversarial noise, and the hyperparameter that controls the distance between $\boldsymbol{\theta}$ and pretrained model ${\boldsymbol{\theta}^{\rm{pre}}}$, respectively. $\cal L$ denotes the output of a three-layer perceptron, $D_j^{\rm{val}}$ and $D_j^{\rm{tr}}$ represent the validation and training datasets on $j^{\rm{th}}$ worker. $\mathcal{Q}(\boldsymbol{P})=\sum\nolimits_{k}r[\max\{ -p_k\!-\!\varepsilon, p_k\!-\! \varepsilon, 0 \}]^2$ is the penalty term for the $L_{\infty}$-ball constraint $||\boldsymbol{P}||_{\infty} \le \varepsilon$ \cite{madry2018towards},  where $p_k$ is the $k^{\rm{th}}$ element in $\boldsymbol{P}$, and $r>0, \varepsilon>0$ are constants. To evaluate the effectiveness of the proposed FPLL, the digit recognition task described in \cite{qian2019robust} is examined. Three benchmark datasets are utilized in this task, i.e., MNIST, USPS, and Fashion MNIST. In the experiments, we set $M=10$, and we utilize the PGD-7 attack \cite{madry2018towards} with $\varepsilon=0.05$. $\cal L$ denotes the output of a three-layer perceptron. 
To ensure consistent formatting, in accordance with \cite{qian2019robust}, we resize images in MNIST and Fashion MNIST to be 16 $\times$ 16. Additionally, we replicate the gray channel in RGB channels to generate the color images. FGSM \cite{goodfellow2014explaining} and PGD-7 \cite{madry2018towards} are implemented to fine-tune the pre-trained model in a non-federated and adversarially robust manner. Since we aim to maintain the performance of the pretraining model while improving the performance and robustness in the fine-tuning task, we employ the average accuracy across pretraining, fine-tuning accuracy, and fine-tuning adversarial robustness as the metric. Comparisons between the proposed FPLL and the SOTA approaches are shown in Fig. \ref{fig:FT_1}-\ref{fig:FT_4}. It is seen from Fig. \ref{fig:FT_1}-\ref{fig:FT_4} that the proposed FPLL exhibits exceptional performance because 1) in comparison with FGSM, PGD, ADBO, FEDNEST, and FMGDA, the proposed FPLL can effectively address the higher-nested optimization problem in a federated manner, rendering it more effective; 2) in contrast to AFTO, the non-linear $L$-cut is used in the proposed FPLL, which has better approximation ability than the linear $\mu$-cut in AFTO.  Additionally, the distributed multi-gradient descent method is used in the proposed FPLL, which can efficiently address the PLL problems with potentially conflicting multi-objectives. In addition, to evaluate the efficiency of the proposed FPLL, we compare it with the state-of-the-art distributed trilevel optimization method AFTO in terms of the number of communication rounds. \textcolor{black}{In the experiment, we maintain comparable compute budgets for the two methods. The key differences are that FPLL employs the more effective non-linear $L$-cut relaxation and utilizes multi-gradient descent tailored for multi-objective optimization.} As illustrated in Fig. \ref{fig:compare_afto}, FPLL converges with fewer communication rounds while achieving superior performance.

\section{Conclusion and Future Work}
Existing multi-level optimization works focus on bilevel and trilevel optimization while multi-level optimization problems involving more than three levels, referred to as PLL problems, which are much harder to solve, remain less explored. Additionally, extant studies on PLL predominantly put attention on the non-federated setting which may result in privacy breaches. To this end, a Federated Poly-Level Learning (FPLL) framework is proposed in this work to tackle the PLL problems with soft constraints in a federated manner. Within the proposed FPLL, we can flexibly control the trade-off between computational complexity and performance. Furthermore, the iteration complexity for the proposed FPLL to achieve the $\epsilon$-Pareto stationary point is theoretically derived. 

\textcolor{black}{As a direction for future work, it would be interesting to extend FPLL to explicitly bounded feasible domains by incorporating projection-based updates, thereby guaranteeing the boundedness of the iterates directly through the algorithmic design. This extension would also require developing a corresponding convergence analysis that accounts for the projection operations and their interaction with multiple local updates in federated settings.}


%

\appendices
\section{Detailed Proofs for Propositions \ref{prop:1} and \ref{prop:2}}
\label{appendix:prop}

\noindent\textit{Proof of Proposition \ref{prop:1}:}

First, let $\mathcal{G}_i$ denote the feasible region with regard to the constraint ${g_i}({\boldsymbol{x}_1}, \cdots ,{\boldsymbol{x}_n}) \le \varepsilon_i $. {\color{black}According to the property of a function with $L$-Lipschitz continuous gradient}, for any point $({\boldsymbol{x}_1}, \cdots ,{\boldsymbol{x}_n}) \in \mathcal{G}_i$, $\forall i $, we have
\begin{equation}
\label{eq:1_6_34}
\begin{array}{l}
    {g_i}({\boldsymbol{x}_1^{t+1}},\! \cdots \!,{\boldsymbol{x}_n^{t+1}})\!+\! {\left[  \begin{array}{l}
\frac{{\partial {g_i}(\boldsymbol{x}_1^{t + 1}, \cdots  ,\boldsymbol{x}_n^{t + 1})}}{{\partial {\boldsymbol{x}_1}}}\\
\qquad \quad \; \vdots \\
 \frac{{\partial {g_i}(\boldsymbol{x}_1^{t + 1}, \cdots  ,\boldsymbol{x}_n^{t + 1})}}{{\partial {\boldsymbol{x}_n}}}
\end{array}  \right]^{\top}}\!  \left[ \begin{array}{l}
{\boldsymbol{x}_1}  - \boldsymbol{x}_1^{t + 1}\\
\quad \; \;  \vdots \\
{\boldsymbol{x}_n} - \boldsymbol{x}_n^{t + 1}
\end{array}  \right]
\vspace{1mm} \\

\le {g_i}({\boldsymbol{x}_1}, \cdots ,{\boldsymbol{x}_n}) + \frac{L}{2} || \left[ \begin{array}{l}
{\boldsymbol{x}_1}  - \boldsymbol{x}_1^{t + 1} \\
\quad \; \;  \vdots \\
{\boldsymbol{x}_n} - \boldsymbol{x}_n^{t + 1}
\end{array}  \right]  ||^2
\vspace{1mm} \\
\le \varepsilon_i +  \frac{L}{2}\sum\nolimits_{i \in [n]} {||{\boldsymbol{x}_i} - \boldsymbol{x}_i^{t + 1}|{|^2}}, 
\end{array}
\end{equation}
where $(\boldsymbol{x}_1^{t + 1}, \cdots  ,\boldsymbol{x}_n^{t + 1})$ represents the point in $(t+1)^{\rm{th}}$ iteration. In accordance with Eq. (\ref{eq:1_6_34}), we can infer that any point $({\boldsymbol{x}_1}, \cdots ,{\boldsymbol{x}_n}) \in \mathcal{G}_i$ must belong to $\boldsymbol{\mathcal{Q}}_i^{t+1}$, which is composed of $L$-cuts.  Thus, we have that the feasible region of constraint ${g_i}({\boldsymbol{x}_1}, \cdots ,{\boldsymbol{x}_n}) \le \varepsilon_i$ is a subset of $\boldsymbol{\mathcal{Q}}_i^{t+1}$. Because ${g_i}({\boldsymbol{x}_1}, \cdots ,{\boldsymbol{x}_n}) \le \varepsilon_i$ is a relaxed version of the constraint ${g_i}({\boldsymbol{x}_1}, \cdots ,{\boldsymbol{x}_n}) = 0$, we can conclude that the feasible region of constraint ${g_i}({\boldsymbol{x}_1}, \cdots ,{\boldsymbol{x}_n}) = 0$ is a subset of $\boldsymbol{\mathcal{Q}}_i^{t+1}=\{\sum\limits_{r \in [n]} \! \sum\limits_{k \in [d_r]} \! { {a_{r, k, l}^{(i)} x_{r,k}^2} \!+\! {b_{r, k, l}^{(i)} x_{r,k}} } - c_l^{(i)} \le  \varepsilon_i, l \! \in \! [|\boldsymbol{\mathcal{Q}}_{i}^{t+1}|] \}$.


\vspace{3mm}

\noindent\textit{Proof of Proposition \ref{prop:2}:}

For any infeasible solution $({\boldsymbol{x}_1'}, \cdots ,{\boldsymbol{x}_n'})$ with respect to constraint ${g_i}({\boldsymbol{x}_1}, \cdots ,{\boldsymbol{x}_n}) \le \varepsilon_i$, according to Eq. (\ref{eq:1_6_34}), the new generated $L$-cut can be expressed as
\begin{equation}
\label{eq:3_25_43}
\begin{array}{l}
 {\left[  \begin{array}{l}
 \frac{{\partial {g_i}(\boldsymbol{x}_1', \cdots   ,\boldsymbol{x}_n')}}{{\partial {\boldsymbol{x}_1}}}\\
\qquad \quad \; \vdots \\
 \frac{{\partial {g_i}(\boldsymbol{x}_1', \cdots  ,\boldsymbol{x}_n')}}{{\partial {\boldsymbol{x}_n}}}
\end{array}  \right]^{\top}}  \left[ \begin{array}{l}
{\boldsymbol{x}_1}  - \boldsymbol{x}_1'\\
\quad \; \;  \vdots \\
{\boldsymbol{x}_n} - \boldsymbol{x}_n'
\end{array}  \right] + g_i(\boldsymbol{x}_1',  \cdots  ,\boldsymbol{x}_n') \vspace{1mm}\\- \frac{L}{2}\sum\nolimits_{i \in [n]} {||{\boldsymbol{x}_i} - \boldsymbol{x}_i'|{|^2}}
 \le \varepsilon_i.
\end{array}
\end{equation}

For any point $({\boldsymbol{x}_1}, \cdots ,{\boldsymbol{x}_n})$ that satisfies ${g_i}({\boldsymbol{x}_1}, \cdots ,{\boldsymbol{x}_n}) \le \varepsilon_i $, we can obtain the following inequality based on the $L$-smoothness:
\begin{equation}
\label{eq:3_23_43}
\begin{array}{l}
 {\left[  \begin{array}{l}
 \frac{{\partial {g_i}(\boldsymbol{x}_1', \cdots   ,\boldsymbol{x}_n')}}{{\partial {\boldsymbol{x}_1}}}\\
\qquad \quad \; \vdots \\
 \frac{{\partial {g_i}(\boldsymbol{x}_1', \cdots  ,\boldsymbol{x}_n')}}{{\partial {\boldsymbol{x}_n}}}
\end{array}  \right]^{\top}}  \left[ \begin{array}{l}
{\boldsymbol{x}_1}  - \boldsymbol{x}_1'\\
\quad \; \;  \vdots \\
{\boldsymbol{x}_n} - \boldsymbol{x}_n'
\end{array}  \right] \!+\! g_i(\boldsymbol{x}_1',  \cdots  ,\boldsymbol{x}_n') \vspace{1mm} \\- \frac{L}{2}\sum\nolimits_{i \in [n]} {||{\boldsymbol{x}_i} - \boldsymbol{x}_i'|{|^2}}
\! \le \! {g_i}(\boldsymbol{x}_1, \cdots  ,\boldsymbol{x}_n).
\end{array}
\end{equation}

By combining Eq. (\ref{eq:3_23_43}) with ${g_i}({\boldsymbol{x}_1}, \cdots ,{\boldsymbol{x}_n}) \le \varepsilon_i $, we can obtain
\begin{equation}
\label{eq:3_23_44}
\begin{array}{l}
 {\left[  \begin{array}{l}
 \frac{{\partial {g_i}(\boldsymbol{x}_1', \cdots   ,\boldsymbol{x}_n')}}{{\partial {\boldsymbol{x}_1}}}\\
\qquad \quad \; \vdots \\
 \frac{{\partial {g_i}(\boldsymbol{x}_1', \cdots  ,\boldsymbol{x}_n')}}{{\partial {\boldsymbol{x}_n}}}
\end{array}  \right]^{\top}}  \left[ \begin{array}{l}
{\boldsymbol{x}_1}  - \boldsymbol{x}_1'\\
\quad \; \;  \vdots \\
{\boldsymbol{x}_n} - \boldsymbol{x}_n'
\end{array}  \right] + g_i(\boldsymbol{x}_1',  \cdots  ,\boldsymbol{x}_n') \vspace{1mm} \\- \frac{L}{2}\sum\nolimits_{i \in [n]} {||{\boldsymbol{x}_i} - \boldsymbol{x}_i'|{|^2}}
 \le \varepsilon_i.
\end{array}
\end{equation}
For the infeasible solution $({\boldsymbol{x}_1'}, \cdots ,{\boldsymbol{x}_n'})$ which violates the constraint ${g_i}({\boldsymbol{x}_1}, \cdots ,{\boldsymbol{x}_n}) \le \varepsilon_i $, i.e., ${g_i}({\boldsymbol{x}_1'}, \cdots ,{\boldsymbol{x}_n'}) > \varepsilon_i$, we can obtain
\begin{equation}
\label{eq:3_23_45}
\begin{array}{l}
 {\left[  \begin{array}{l}
 \frac{{\partial {g_i}(\boldsymbol{x}_1', \cdots   ,\boldsymbol{x}_n')}}{{\partial {\boldsymbol{x}_1}}}\\
\qquad \quad \; \vdots \\
 \frac{{\partial {g_i}(\boldsymbol{x}_1', \cdots  ,\boldsymbol{x}_n')}}{{\partial {\boldsymbol{x}_n}}}
\end{array}  \right]^{\top}}  \left[ \begin{array}{l}
{\boldsymbol{x}_1'}  - \boldsymbol{x}_1'\\
\quad \; \;  \vdots \\
{\boldsymbol{x}_n'} - \boldsymbol{x}_n'
\end{array}  \right] + g_i(\boldsymbol{x}_1',  \cdots  ,\boldsymbol{x}_n') \vspace{1mm}\\- \frac{L}{2}\sum\nolimits_{i \in [n]} {||{\boldsymbol{x}_i'} - \boldsymbol{x}_i'|{|^2}}
\vspace{1mm}\\ = {g_i}(\boldsymbol{x}_1', \cdots  ,\boldsymbol{x}_n')
\\ >\varepsilon_i.
\end{array}
\end{equation}
According to Eq. (\ref{eq:3_25_43}), (\ref{eq:3_23_44}), and (\ref{eq:3_23_45}), we can deduce that the newly generated $L$-cut can separate the infeasible solution $(\boldsymbol{x}_1', \cdots   ,\boldsymbol{x}_n')$ from the feasible region in terms of constraint ${g_i}(\boldsymbol{x}_1, \cdots  ,\boldsymbol{x}_n)\le \varepsilon_i $. Consequently, $\boldsymbol{\mathcal{Q}}_i^{t+1}$ converges monotonically with the incremental addition of $L$-cuts, implying that:
\begin{equation}
    \mathcal{G}_i \subseteq \boldsymbol{\mathcal{Q}}_i^{t+1} \subseteq \boldsymbol{\mathcal{Q}}_i^{t}  \subseteq  \cdots  \subseteq \boldsymbol{\mathcal{Q}}_i^{0}  , \forall i.
\end{equation}


\section{Detailed Proofs for Theorem \ref{theorem:1}}\label{appendix:proof-theorem}
In this part, we analyze the upper bound of the iteration complexity for the proposed FPLL to attain the $\epsilon$-Pareto stationary point.  First, based on updates of variables in the distributed multi-gradient descent method, i.e., $\boldsymbol{x}_i^{t + 1}=\boldsymbol{x}_i^{t}-\alpha \boldsymbol{s}_i^t, \forall i \in [n]$, we can derive
\begin{equation}
    \begin{array}{l}
\frac{1}{M}\sum\limits_{j =1}^M {{q_{1,j}}(\boldsymbol{x}_1^{t + 1}, \cdots ,\boldsymbol{x}_n^{t + 1})}  - \frac{1}{M}\sum\limits_{j =1}^M {{q_{1,j}}(\boldsymbol{x}_1^t, \cdots ,\boldsymbol{x}_n^t)} \\

= \frac{1}{M} \!\sum\limits_{j =1}^M {{q_{1,j}}(\boldsymbol{x}_1^t - \alpha \boldsymbol{s}_1^t, \cdots ,\boldsymbol{x}_n^{t} - \alpha \boldsymbol{s}_n^t)}  \!-\! {{q_{1,j}}(\boldsymbol{x}_1^t, \cdots ,\boldsymbol{x}_n^t)} .
\end{array}
\end{equation}

According to the $L$-smoothness in Assumption \ref{assum:1}, for any $t>K_1$, we can obtain
\begin{equation}
\label{eq:12_19_22}
    \begin{array}{l}
\frac{1}{M}\sum\nolimits_{j =1}^M {{q_{1,j}}(\boldsymbol{x}_1^{t + 1}, \cdots ,\boldsymbol{x}_n^{t + 1})}  - \frac{1}{M}\sum\nolimits_{j =1}^M {{q_{1,j}}(\boldsymbol{x}_1^t, \cdots ,\boldsymbol{x}_n^t)} \\

= \frac{1}{M}\sum\limits_{j =1}^M {{q_{1,j}}(\boldsymbol{x}_1^t - \alpha \boldsymbol{s}_1^t, \cdots ,\boldsymbol{x}_n^{t} - \alpha \boldsymbol{s}_n^t)}  - {{q_{1,j}}(\boldsymbol{x}_1^t, \cdots ,\boldsymbol{x}_n^t)}  \\
 
 \le  - \frac{\alpha }{M}{\sum\nolimits_{j =1}^M {\nabla {q_{1,j}}(\boldsymbol{x}_1^t, \cdots ,\boldsymbol{x}_n^{t})} ^{\top}}{\boldsymbol{S}^t} + \frac{{L{\alpha ^2}}}{2}||{\boldsymbol{S}^t}|{|^2}.
\end{array}
\end{equation}

According to the equality $\boldsymbol{a}^{\top}\boldsymbol{b}=\frac{1}{2}(||\boldsymbol{a}||^2 + ||\boldsymbol{b}||^2 - ||\boldsymbol{a}-\boldsymbol{b}||^2)$, it follows that
\begin{equation}
\label{eq:12_19_23}
    \begin{array}{l}
 - \frac{\alpha }{M}{\sum\nolimits_{j =1}^M {\nabla {q_{1,j}}(\boldsymbol{x}_1^t, \cdots ,\boldsymbol{x}_n^t)} ^{\top}}{\boldsymbol{S}^t} \\
 
 =  - \frac{\alpha }{2}( ||\frac{1}{M}\sum\nolimits_{j =1}^M {\nabla {q_{1,j}}(\boldsymbol{x}_1^t, \cdots ,\boldsymbol{x}_n^t)} |{|^2} + ||{\boldsymbol{S}^t}|{|^2} \\ 
 
 - ||\frac{1}{M}\sum\nolimits_{j =1}^M {\nabla {q_{1,j}}(\boldsymbol{x}_1^t, \cdots ,\boldsymbol{x}_n^t)}  - {\boldsymbol{S}^t}|{|^2} ).
\end{array}
\end{equation}

Though integrating Eq. (\ref{eq:12_19_22}) with Eq. (\ref{eq:12_19_23}), we can obtain
\begin{equation}
\label{eq:2_1_47}
    \begin{array}{l}
\frac{1}{M}\sum\limits_{j =1}^M {{q_{1,j}}(\boldsymbol{x}_1^{t + 1}, \cdots ,\boldsymbol{x}_n^{t + 1})}  - \frac{1}{M}\sum\limits_{j =1}^M {{q_{1,j}}(\boldsymbol{x}_1^t, \cdots ,\boldsymbol{x}_n^t)} \\

 \le  - \frac{\alpha }{2}( ||\frac{1}{M}\sum\nolimits_{j =1}^M {\nabla {q_{1,j}}(\boldsymbol{x}_1^t, \cdots ,\boldsymbol{x}_n^t)} |{|^2} + ||{\boldsymbol{S}^t}|{|^2}
 \\ - ||\frac{1}{M}\sum\nolimits_{j =1}^M {\nabla {q_{1,j}}(\boldsymbol{x}_1^t, \cdots ,\boldsymbol{x}_n^t)}  - {\boldsymbol{S}^t}|{|^2} )  + \frac{{L{\alpha ^2}}}{2}||{\boldsymbol{S}^t}|{|^2} \\
 
\le  - \frac{\alpha }{2}( ||\frac{1}{M}\sum\nolimits_{j =1}^M {\nabla {q_{1,j}}(\boldsymbol{x}_1^t, \cdots ,\boldsymbol{x}_n^t)} |{|^2} \\- ||\frac{1}{M}\sum\nolimits_{j =1}^M {\nabla {q_{1,j}}(\boldsymbol{x}_1^t, \cdots ,\boldsymbol{x}_n^t)}  - {\boldsymbol{S}^t}|{|^2} ) . 
\end{array}
\end{equation}

In the last inequality in Eq. (\ref{eq:2_1_47}), we utilize the setting that $\alpha \le \frac{1}{L}$. 
And according to the constraint in Eq. (\ref{eq:12_8_eq4}), i.e.,
\begin{equation}
\begin{array}{l}
    ||\frac{1}{M}\sum\nolimits_{j =1}^M {\nabla {q_{1,j}}(\boldsymbol{x}_1^t, \cdots ,\boldsymbol{x}_n^t)}  - {\boldsymbol{S}^t}|{|^2} \\  \le {c^2}||\frac{1}{M}\sum\nolimits_{j =1}^M {\nabla {q_{1,j}}(\boldsymbol{x}_1^t, \cdots ,\boldsymbol{x}_n^t)} |{|^2},
\end{array}
\end{equation}
we can derive the following inequality:
\begin{equation}
\label{eq:12_19_26}
    \begin{array}{l}
\frac{1}{M}\sum\nolimits_{j =1}^M {{q_{1,j}}(\boldsymbol{x}_1^{t + 1}, \cdots ,\boldsymbol{x}_n^{t + 1})}  - \frac{1}{M}\sum\nolimits_{j =1}^M {{q_{1,j}}(\boldsymbol{x}_1^t, \cdots ,\boldsymbol{x}_n^t)} \\

 \le  \frac{\alpha }{2} \left( {{c^2} - 1} \right)||\frac{1}{M}\sum\limits_{j =1}^M {\nabla {q_{1,j}}(\boldsymbol{x}_1^t, \cdots ,\boldsymbol{x}_n^t)} |{|^2}.
\end{array}
\end{equation}

By summing up the inequality in Eq. (\ref{eq:12_19_26}) over iterations from $t=K_1$ to $T(\epsilon)$, we can derive the inequality below.
\begin{equation}
\label{eq:12_19_27}
    \begin{array}{l}
\frac{1}{M}\sum\limits_{j =1}^M {{q_{1,j}}(\boldsymbol{x}_1^{T(\epsilon )}, \cdots ,\boldsymbol{x}_n^{T(\epsilon )})}  - \frac{1}{M}\sum\limits_{j =1}^M {{q_{1,j}}(x_1^{K_1}, \cdots ,x_n^{K_1})} \\

 \le \frac{\alpha }{2} \left( {{c^2} - 1} \right)\sum\limits_{t = {K_1}}^{T(\epsilon )} {||\frac{1}{M}\sum\limits_{j =1}^M {\nabla {q_{1,j}}(\boldsymbol{x}_1^t, \cdots ,\boldsymbol{x}_n^t)} |{|^2}} .
\end{array}
\end{equation}

Since the distributed multi-gradient descent algorithm aims to find the descent direction for all objectives at every iteration, and all variables are bounded, we can obtain that $\frac{1}{M}\sum\nolimits_j  {{q_{1,j}}(x_1^{K_1}, \cdots ,x_n^{K_1})} $ is upper bounded.  By combining Eq. (\ref{eq:12_19_27}) with Assumption \ref{assum:2}, i.e., $q_{1,j}^* > -\infty$ and ${1 - {c^2}}>0$, we can obtain
\begin{equation}
\label{eq:12_26_37}
    \begin{array}{l}
\mathop {\min }\limits_t \mathop {\min }\limits_{\boldsymbol{p} \in \Delta } ||\sum\nolimits_{j =1}^M {{p_j}\nabla {q_{1,j}}({\boldsymbol{x}_1^t}, \cdots ,{\boldsymbol{x}_n^t})} |{|^2} \\

 \le \mathop {\min }\limits_t ||\frac{1}{M}\sum\nolimits_{j =1}^M {\nabla {q_{1,j}}(\boldsymbol{x}_1^t, \cdots ,\boldsymbol{x}_n^t)} |{|^2} \\
 
 \le \frac{1}{{T(\varepsilon ) - K_1 + 1}}\sum\limits_{t = K_1}^{T(\epsilon )} {||\frac{1}{M}\sum\limits_{j =1}^M {\nabla {q_{1,j}}(\boldsymbol{x}_1^t, \cdots ,\boldsymbol{x}_n^t)} |{|^2}} \\

  \le \frac{ \frac{1}{M}\sum\limits_{j =1}^M {{q_{1,j}}(x_1^{K_1}, \cdots ,x_n^{K_1})}  -  \frac{1}{M}\sum\limits_{j =1}^M {{q_{1,j}}(\boldsymbol{x}_1^{T(\epsilon )}, \cdots ,\boldsymbol{x}_n^{T(\epsilon )})}  }{{\frac{\alpha }{2}\left( {1 - {c^2}} \right)\left( {T(\epsilon ) -K_1 + 1} \right)}}   \\

 \le \frac{ \frac{1}{M}\sum\limits_{j =1}^M \max\limits_{k \in [K_1]} {{q_{1,j}}(x_1^{k}, \cdots ,x_n^{k})}  -  q_{1,j}^*  }{{\frac{\alpha }{2}\left( {1 - {c^2}} \right)\left( {T(\epsilon ) -K_1 + 1} \right)}} .
\end{array}
\end{equation}

By combining Eq. (\ref{eq:12_26_37}) with the definitions of $\epsilon$-Pareto stationary point and $T(\epsilon)$, we can conclude
\begin{equation}
\begin{array}{l}
T(\epsilon)\sim  \frac{( \frac{1}{M}\sum\limits_{j =1}^M \max\limits_{k \in [K_1]} {{q_{1,j}}(x_1^{k}, \cdots ,x_n^{k})}  -  q_{1,j}^*  )}{{\frac{\alpha }{2}\left( {1 - {c^2}} \right) \epsilon}} + K_1 - 1 .
\end{array}
\end{equation}

\section{Detailed Proofs for Theorem \ref{theorem:2}}
\label{appendix:proof-theorem2}

In this work, the communication complexity is composed of two primary parts: (i) the cost of transmitting gradients and model parameters at each iteration, and (ii) the communication overhead associated with updating the $L$-cuts.

(i) In each iteration, the workers transmit the gradients $\nabla {q_{1,j}}(\boldsymbol{x}_1^t,\! \cdots \!,\boldsymbol{x}_n^t) \in \mathbb{R}^{\sum_i d_i}$  to the master, and the master broadcasts the updated global variables $\boldsymbol{x}_i^{t + 1} \in \mathbb{R}^{d_i}, \forall i \in [n]$ to the workers. Thus, the communication complexity in the entire iterative process is
\begin{equation}
\label{eq:1_13_38}
\begin{array}{l}
32(2\sum\nolimits_{i=1}^nd_i)MT(\epsilon).
\end{array}
\end{equation}

(ii) In the proposed method, the $L$-cuts will be updated every $\tau$ iterations when $t<K_1$. Thus, the number of communication rounds required for updating the $L$-cuts is upper bounded by $\lfloor \frac{K_1}{\tau} \rfloor$. According to the discussion in Appendix \ref{appendix:L_cut}, when updating the $m^{\rm{th}}$ level $L$-cut for any $m \in \{2, \dots, n\}$, each worker transmits the local accumulated gradient $\Delta_{m,j} \in \mathbb{R}^{d_m}$ to the master, and the master broadcasts the updated $2\boldsymbol{\delta}_m^{t+1} \in \mathbb{R}^{d_m}$ to the workers. Then, the local derivative-vector product $\boldsymbol{h}_{m,j}^{t+1} \in \mathbb{R}^{\sum\nolimits_{i=1}^{n}d_i}$ is communicated to the master, and the new $m^{\rm{th}}$ level $L$-cut is generated. Subsequently, the newly generated $L$-cut parameters $a_{r, k, l}^{(m)}, b_{r, k, l}^{(m)}, c_l^{(m)}$, for $r \in [n]$ and $k \in [d_r]$, are transmitted to the workers. Summing these communication costs from the $2^{\rm{nd}}$ to the $n^{\rm{th}}$ layer, the overall communication overhead associated with updating the $L$-cuts over the entire iterative process is 
\begin{equation}
\label{eq:1_13_39}
\begin{array}{l}
32\sum\nolimits_{m=2}^n(2d_m+3\sum\nolimits_{i=1}^nd_i+1)M\lfloor \frac{K_1}{\tau} \rfloor \\
\le 32 (3\sum\nolimits_{i=1}^n d_i +1) n M \lfloor \frac{K_1}{\tau} \rfloor.
\end{array}
\end{equation}

Therefore, based on Eq. (\ref{eq:1_13_38}) and Eq. (\ref{eq:1_13_39}), the communication complexity of the proposed method to reach an $\epsilon$-Pareto stationary point is upper bounded by
\begin{equation}
\begin{array}{l}
\! 32(2\sum\nolimits_{i=1}^nd_i)MT(\epsilon)\!+\!32(3\sum\nolimits_{i=1}^nd_i+1)nM\lfloor \frac{K_1}{\tau} \rfloor.
\end{array}
\end{equation}

\section{Distributed Multi-gradient Descent Method} \label{appendix:multi-gradient}
In this section, the details of obtaining the search direction in the introduced distributed multi-gradient descent method are provided. By combining the first-order Taylor approximation with Eq. (\ref{eq:4_28_13}), we have
\begin{equation}
\begin{array}{l}
    \mathop {\max }\limits_{j \in [M]} \left\{ {\frac{1}{\alpha }\! \left( {{q_{1,j}}({\boldsymbol{x}_1} - {\boldsymbol{s}_1}, \! \cdots \! ,{\boldsymbol{x}_n} - {\boldsymbol{s}_n}) - {q_{1,j}}({\boldsymbol{x}_1}, \! \cdots \! ,{\boldsymbol{x}_n})} \right)} \right\} \\
 \approx  - \mathop {\min }\limits_{j \in [M]} \left\langle {\nabla {q_{1,j}({\boldsymbol{x}_1}, \! \cdots \! ,{\boldsymbol{x}_n})},\boldsymbol{S}} \right\rangle .
\end{array}
\end{equation}

In $(t+1)^{\rm{th}}$ iteration, the obtaining of $\boldsymbol{S}$ can be framed as a constrained optimization problem as follows:
\begin{equation}
\label{eq:12_8_eq4}
    \begin{array}{l}
\mathop {\max }\limits_{\boldsymbol{S}} \mathop {\min }\limits_{j \in [M]} \left\langle {\nabla {q_{1,j}}({\boldsymbol{x}_1^t}, \cdots ,{\boldsymbol{x}_n^t}),\boldsymbol{S}} \right\rangle \\
{\rm{s}}{\rm{.t}}{\rm{. }}\;||\boldsymbol{S} - \frac{1}{M}\sum\nolimits_{j = 1}^M {\nabla {q_{1,j}}}({\boldsymbol{x}_1^t}, \cdots ,{\boldsymbol{x}_n^t}) ||
\\ \qquad   \le c||\frac{1}{M}\sum\nolimits_{j = 1}^M {\nabla {q_{1,j}}}({\boldsymbol{x}_1^t}, \cdots ,{\boldsymbol{x}_n^t}) || 
\end{array}
\end{equation}

In Eq. (\ref{eq:12_8_eq4}), $0<c<1$ is a pre-set constant which constrains that the search direction $\boldsymbol{S}$ is around the averaged gradient $\frac{1}{M}\sum\limits_{j = 1}^M {\nabla {q_{1,j}}}({\boldsymbol{x}_1^t}, \cdots ,{\boldsymbol{x}_n^t})$. Directly solving the optimization problem in Eq. (\ref{eq:12_8_eq4}) is computationally complex because the $\boldsymbol{S} \in \mathbb{R}^{\sum_i d_i}$ is usually a high-dimensional vector. To this end, the dual problem is considered. First we have that $\mathop {\min }\limits_{j \in [M]} \left\langle {\nabla {q_{1,j}}({\boldsymbol{x}_1^t}, \cdots ,{\boldsymbol{x}_n^t}),\boldsymbol{S}} \right\rangle  = \mathop {\min }\limits_{\boldsymbol{p} \in \Delta } \left\langle {\sum\nolimits_{j =1}^M {{p_j}\nabla {q_{1,j}}({\boldsymbol{x}_1^t}, \cdots ,{\boldsymbol{x}_n^t})} ,\boldsymbol{S}} \right\rangle $ \cite{gu2023min}, where $\Delta=\{\boldsymbol{p}| 0 \le p_j\le 1, \forall j, \sum\nolimits_{j =1}^M p_j = 1 \}$ is the simplex. The Lagrangian function can be expressed as
\begin{equation}
\begin{array}{l}
\mathop {\max }\limits_{\boldsymbol{S}} \mathop {\min }\limits_{\boldsymbol{p} \in \Delta ,\phi  \ge 0} \left\langle {\sum\nolimits_{j =1}^M {{p_j}\nabla {q_{1,j}}({\boldsymbol{x}_1^t}, \cdots ,{\boldsymbol{x}_n^t})} ,\boldsymbol{S}} \right\rangle \\
- \frac{\phi }{2}||\boldsymbol{S} - \frac{1}{M}\sum\nolimits_{j = 1}^M {\nabla {q_{1,j}}}({\boldsymbol{x}_1^t}, \cdots ,{\boldsymbol{x}_n^t}) |{|^2} \\+\frac{\phi {c^2}}{2} ||\frac{1}{M}\sum\nolimits_{j = 1}^M {\nabla {q_{1,j}}}({\boldsymbol{x}_1^t}, \cdots ,{\boldsymbol{x}_n^t}) |{|^2}
\\
{\rm{var.}} \qquad \qquad \qquad \quad \boldsymbol{S},  \boldsymbol{p}, \phi,
\end{array}
\end{equation}
where $\phi \in \mathbb{R}^1$ represents the dual variable for the inequality constraint. Following \cite{liu2021conflict}, the min and max are switched without altering the solution, as guided by Slater’s condition:
\begin{equation}
\label{eq:12_8_11}
\begin{array}{l}
\mathop {\min }\limits_{\boldsymbol{p} \in \Delta ,\phi  \ge 0} \mathop {\max }\limits_{\boldsymbol{S}} \left\langle {\sum\nolimits_{j =1}^M {{p_j}\nabla {q_{1,j}}({\boldsymbol{x}_1^t}, \cdots ,{\boldsymbol{x}_n^t})} ,\boldsymbol{S}} \right\rangle \\
- \frac{\phi }{2}||\boldsymbol{S} - \frac{1}{M}\sum\nolimits_{j = 1}^M {\nabla {q_{1,j}}}({\boldsymbol{x}_1^t}, \cdots ,{\boldsymbol{x}_n^t}) |{|^2} \\+\frac{\phi {c^2}}{2} ||\frac{1}{M}\sum\nolimits_{j = 1}^M {\nabla {q_{1,j}}}({\boldsymbol{x}_1^t}, \cdots ,{\boldsymbol{x}_n^t}) |{|^2}
\\
{\rm{var.}} \qquad \qquad \qquad \quad \boldsymbol{S},  \boldsymbol{p}, \phi.
\end{array}
\end{equation}

By fixing the variables $\boldsymbol{p}$ and $\phi$, we can obtain the optimal $\boldsymbol{S}^*$ as
\begin{equation}
\label{eq:12_8_12}
\begin{array}{l}
    \boldsymbol{S}^* = \frac{{\sum\limits_{j =1}^M {{p_j}\nabla {q_{1,j}} ({\boldsymbol{x}_1^t}, \! \cdots \!,{\boldsymbol{x}_n^t})} }}{\phi } + \frac{1}{M}\sum\limits_{j = 1}^M {\nabla {q_{1,j}}({\boldsymbol{x}_1^t}, \! \cdots \!,{\boldsymbol{x}_n^t})} .
\end{array}
\end{equation}

By combining the optimal $\boldsymbol{S}^*$ from Eq. (\ref{eq:12_8_12}) with Eq. (\ref{eq:12_8_11}), we arrive at the following optimization problem involving variables $\boldsymbol{p}, \phi$:
\begin{equation}
\label{eq:12_8_13}
\begin{array}{l}
\!\mathop {\min }\limits_{\boldsymbol{p} \in \Delta ,\phi  \ge 0}\! \left\langle {\sum\limits_{j =1}^M\! {{p_j}\nabla {q_{1,j}}({\boldsymbol{x}_1^t}, \!\cdots \!,{\boldsymbol{x}_n^t})} ,\frac{1}{M} \!\sum\limits_{j = 1}^M \!{\nabla {q_{1,j}}({\boldsymbol{x}_1^t}, \!\cdots \!,{\boldsymbol{x}_n^t})} } \right\rangle 
\\ + \frac{1}{{2\phi }}||\sum\nolimits_{j =1}^M {{p_j}\nabla {q_{1,j}}({\boldsymbol{x}_1^t}, \cdots ,{\boldsymbol{x}_n^t})} |{|^2} 
\\ + \frac{{\phi {c^2}}}{2}||\frac{1}{M}\sum\nolimits_{j = 1}^M {\nabla {q_{1,j}}({\boldsymbol{x}_1^t}, \cdots ,{\boldsymbol{x}_n^t})} |{|^2}
\\
{\rm{var.}} \qquad \qquad \qquad \qquad \qquad  \boldsymbol{p}, \phi.
\end{array}
\end{equation}

Then, we can determine the optimal $\phi^*$ for the problem stated in Eq. (\ref{eq:12_8_13}) by fixing the variable $\boldsymbol{p}$ in Eq. (\ref{eq:12_8_13}), which is given by:
\begin{equation}
    \phi^*  = \frac{1}{c}\frac{{||\sum\nolimits_{j =1}^M {{p_j}\nabla {q_{1,j}}({\boldsymbol{x}_1^t}, \cdots ,{\boldsymbol{x}_n^t})} ||}}{{||\frac{1}{M}\sum\nolimits_{j = 1}^M {\nabla {q_{1,j}}}({\boldsymbol{x}_1^t}, \cdots ,{\boldsymbol{x}_n^t}) ||}}.
\end{equation}

Substituting the optimal $\phi^*$ to Eq. (\ref{eq:12_8_13}), the optimization problem is transformed as the following form with respect to variable $\boldsymbol{p}$:
\begin{equation}
\label{eq:12_8_15}
\begin{array}{l}
\mathop {\min }\limits_{\boldsymbol{p} \in \Delta } \left\langle {\sum\limits_{j =1}^M\! {{p_j}\nabla {q_{1,j}}({\boldsymbol{x}_1^t}, \!\cdots \!,{\boldsymbol{x}_n^t})} ,\frac{1}{M}\!\sum\limits_{j = 1}^M\! {\nabla {q_{1,j}}({\boldsymbol{x}_1^t}, \!\cdots \!,{\boldsymbol{x}_n^t})} } \right\rangle \\

+ c||\frac{1}{M}\!\sum\limits_{j = 1}^M\! {\nabla {q_{1,j}}({\boldsymbol{x}_1^t}, \!\cdots \!,{\boldsymbol{x}_n^t})} || \cdot ||\sum\limits_{j =1}^M\! {{p_j}\nabla {q_{1,j}}({\boldsymbol{x}_1^t}, \!\cdots \!,{\boldsymbol{x}_n^t})} ||
\\
{\rm{var.}} \qquad \qquad \qquad \qquad  \boldsymbol{p}.
\end{array}
\end{equation}
By solving the problem in Eq. (\ref{eq:12_8_15}), we can derive the optimal solution $\boldsymbol{p}^*$. Combining this result with Eq. (\ref{eq:12_8_12}), we can determine the search direction in the $(t+1)^{\rm{th}}$ iteration as follows:
\begin{equation}
\label{eq:1_31_39}
\begin{array}{l}
{\boldsymbol{S}^{t + 1}} = \frac{{\sum\limits_{j =1}^M {p_j^*\nabla {q_{1,j}}({\boldsymbol{x}_1^t}, \!\cdots \!,{\boldsymbol{x}_n^t})} }}{\phi^* } + \frac{1}{M}\!\sum\limits_{j = 1}^M\! {\nabla {q_{1,j}}({\boldsymbol{x}_1^t}, \!\cdots \!,{\boldsymbol{x}_n^t})} .
\end{array}
\end{equation}

\vspace{2mm}

\noindent\textbf{Complexity of Solving Problem (\ref{eq:12_8_eq6}):}

We can reformulate the optimization problem in Eq. (\ref{eq:12_8_eq6}) as a Conic Quadratic Programming (CQP) through the inclusion of an additional variable $t \in \mathbb{R}^1$ as follows:
\begin{equation}
\label{eq:10_16_48}
\begin{array}{l}
\mathop {\min }\limits_{\boldsymbol{p} \in \Delta, t } \left\langle {\sum\limits_{j \in [M]} {{p_j}\nabla {q_{1,j}^t}} ,\frac{1}{M}\sum\limits_{j \in [M]} {\nabla {q_{1,j}^t}} } \right\rangle + t
\vspace{1mm}\\
{\rm{s.t.}} \quad c||\frac{1}{M}\sum\limits_{j \in [M]} {\nabla {q_{1,j}^t}} || \cdot ||\sum\limits_{j \in [M]} {{p_j}\nabla {q_{1,j}^t}} || \le t
\end{array}.
\end{equation}

CQP can be solved effectively by polynomial time interior point method, as discussed in \cite{ben2021lectures}, and the arithmetic complexity of obtaining a $\gamma$-solution in problem (\ref{eq:10_16_48}) is $\mathcal{O}\left(\sqrt{2}((M+1)^3+(M+1)d_1^2)\log(\frac{1}{\gamma})\right)$.

\textcolor{black}{
\section{Computation of $L$-cut Parameters}
\label{appendix:L_cut}
In this section, we discuss the computation of $L$-cut parameters in detail. In the proposed framework, the $L$-cut parameters are updated every $\tau$ iteration ($\tau \ge 1$ is a constant). Taking the $n^{\rm{th}}$ level $L$-cut as an example, during the $(t+1)^{\rm th}$ iteration satisfying $(t+1)\bmod \tau = 0$, the worker $j, j\in [M]$ first updates the local variables using $Q$ steps of gradient descent and then transmits the obtained $\Delta_{n,j}^{t+1} \in \mathbb{R}^{d_n}$, i.e.,
\begin{equation}
\begin{array}{l}
    \Delta_{n,j}^{t+1} = \sum_{q=0}^{Q-1} \nabla_{\boldsymbol{x}_n} f_{n,j}({\boldsymbol{x}_1^{t+1}}, \cdots, {\boldsymbol{x}_{n-1}^{t+1}}, {\boldsymbol{x}_{n,j}^q}),
\end{array}
\end{equation}
to the master. Then, the master can compute:
\begin{equation}
\begin{array}{l}
\boldsymbol{\delta}_n^{t+1} = {\boldsymbol{x}_n^{t+1}} - {\boldsymbol{x}_n^0} + \frac{\eta_{\boldsymbol{x}_n}}{M} \sum_{j=1}^M \Delta_{n,j}^{t+1} .
\end{array}
\end{equation} }

\textcolor{black}{Thus, $g_n({\boldsymbol{x}_1^{t+1}}, \cdots, {\boldsymbol{x}_n^{t+1}}) = ||\boldsymbol{\delta}_n^{t+1}||^2$ is obtained in the master, and the master broadcasts $ 2\boldsymbol{\delta}_n^{t+1} \in \mathbb{R}^{d_n}$ to the workers. The worker $j$ can compute and transmit the local derivative-vector product to the master as follows:
\begin{equation}
\boldsymbol{h}_{n,j}^{t+1}
=
\left.
\nabla_{[\boldsymbol{x}_1,\cdots,\boldsymbol{x}_n]}
\left\langle
\Delta_{n,j}^{t+1},\,
2\boldsymbol{\delta}_n^{t+1}
\right\rangle
\right|_{\boldsymbol{x}_i=\boldsymbol{x}_i^{t+1},\, i\in[n]}
\in \mathbb{R}^{\sum\nolimits_{i=1}^{n}d_i}.
\end{equation} }

\textcolor{black}{Here, $\Delta_{n,j}^{t+1}$ is differentiable with respect to $\boldsymbol{x}_i,i\in[n]$, and $2\boldsymbol{\delta}_n^{t+1}$ is treated as constant when computing the derivative. Therefore, the gradients can be obtained in the master:
\begin{equation}
\begin{array}{l}
\nabla_{[{\boldsymbol{x}_1}, \cdots, {\boldsymbol{x}_n}]} g_n({\boldsymbol{x}_1^{t+1}}, \cdots, {\boldsymbol{x}_n^{t+1}}) \\ =  \frac{\eta_{\boldsymbol{x}_n}}{M} \sum\limits_{j=1}^M \boldsymbol{h}_{n,j}^{t+1}  +  \nabla_{[\boldsymbol{x}_1, \cdots, \boldsymbol{x}_n]} ( 2{\boldsymbol{\delta}_n^{t+1}}^\top \boldsymbol{x}_n ) |_{\boldsymbol{x}_n = \boldsymbol{x}_n^{t+1}} 
\end{array}.
\end{equation} }

\textcolor{black}{Consequently, the new $n^{\rm{th}}$ level $L$-cut and its corresponding parameters can be obtained in the master as follows:
\begin{equation}
\label{eq:5_17_55}
\begin{array}{l}
\sum\limits_{r \in [n]} \! \sum\limits_{k \in [d_r]} \! { {a_{r, k, l}^{(n)} x_{r,k}^2} \!+\! {b_{r, k, l}^{(n)} x_{r,k}} } -c_l^{(n)} \le \varepsilon_n \;\;  {\rm{where}} \\
a_{r, k, l}^{(n)} = -\frac{L}{2},\; b_{r, k, l}^{(n)} = \nabla_{x_{r,k}} g_n({\boldsymbol{x}_1^{t+1}}, \cdots, {\boldsymbol{x}_n^{t+1}}) + L x_{r,k}^{t+1}, \vspace{1mm}\\
c_l^{(n)} = \sum_{i=1}^n \left( \nabla_{{\boldsymbol{x}_i}} g_n({\boldsymbol{x}_1^{t+1}}, \cdots, {\boldsymbol{x}_n^{t+1}})^\top {\boldsymbol{x}_i^{t+1}} \right) \vspace{1mm}\\  \qquad- g_n({\boldsymbol{x}_1^{t+1}}, \cdots, {\boldsymbol{x}_n^{t+1}}) + \frac{L}{2} \sum_{i=1}^n ||{\boldsymbol{x}_i^{t+1}}||^2.
\end{array}
\end{equation} }


\textcolor{black}{Notably, the computation of the $L$-cut parameters avoids raw data transmission, requiring communication between the master and workers only every $\tau$ iterations. Moreover, this derivation naturally generalizes to any intermediate level $i < n$. Upon computing the respective residual function $g_i$ and its gradient $\nabla g_i(\boldsymbol{x}_1^{t+1}, \cdots, \boldsymbol{x}_n^{t+1})$, the $i^{\rm{th}}$ level $L$-cut parameters, i.e., $a_{r, k, l}^{(i)}, b_{r, k, l}^{(i)}$, and $c_l^{(i)}$, can be directly obtained. In practice, the exact smoothness parameter $L$ may not be available. Several methods can be used to estimate the smoothness parameter, including local smoothness estimation in \cite{lan2024projected} and backtracking in \cite{beck2009fast}. For instance, the local smoothness estimator for each level $i$ can be computed as follows:
\begin{equation}
\widehat{L}_i^{t+1} = \sqrt{\frac{\sum_{r=1}^n \|\nabla_{\boldsymbol{x}_r} g_i^{t+1} - \nabla_{\boldsymbol{x}_r} g_i^t\|^2}{\sum_{r=1}^n \|\boldsymbol{x}_r^{t+1} - \boldsymbol{x}_r^t\|^2}},
\end{equation}
where $g_i^{t+1}=g_i(\boldsymbol{x}_1^{t+1},\cdots,\boldsymbol{x}_n^{t+1})$ and we can obtain:
\begin{equation}
  \widehat L^{t+1}=\max_{i\in[n]}\widehat L_i^{t+1}, \;\;
 L=\max\left\{ L,\rho \widehat L^{t+1}
\right\}  ,
\end{equation}
where $\rho\ge1$ is a safety factor. In addition, it is worth noting that the proposed $L$-cut does not require the exact value of $L$, it is sufficient to use an upper bound. A larger smoothness parameter only yields a more conservative cut and does not invalidate the relaxation.}



\ifCLASSOPTIONcaptionsoff
  \newpage
\fi

\bibliographystyle{ieeetr}
\bibliography{ref}

\end{document}